\documentclass[10pt,twocolumn,letterpaper]{article}

\usepackage{wacv}
\usepackage{times}
\usepackage{epsfig}
\usepackage{graphicx}
\usepackage{amsmath}
\usepackage{amssymb}
\usepackage{booktabs}
\usepackage[table,xcdraw]{xcolor}
\definecolor{snsred}{rgb}{0.86, 0.3712, 0.33999999999999997}
\definecolor{snsyellow}{rgb}{0.8287999999999999, 0.86, 0.33999999999999997}
\definecolor{snsgreen}{rgb}{0.33999999999999997, 0.86, 0.3712}
\definecolor{snsazzurro}{rgb}{0.33999999999999997, 0.8287999999999999, 0.86}
\definecolor{snsblue}{rgb}{0.3712, 0.33999999999999997, 0.86}
\definecolor{snsviola}{rgb}{0.86, 0.33999999999999997, 0.8287999999999999}
\usepackage{xcolor}
\usepackage[accsupp]{axessibility}

\def\wacvPaperID{1350} 
\newcommand{\DATASET}{BBBicycles}
\newcommand{\NETWORK}{TransReI3D}
\newcommand{\CLS}{[\textit{cls}]}
\wacvalgorithmstrack   

\wacvfinalcopy 

\ifwacvfinal
\usepackage[breaklinks=true,bookmarks=false]{hyperref}
\else
\usepackage[pagebackref=true,breaklinks=true,colorlinks,bookmarks=false]{hyperref}
\fi
\usepackage{subcaption}
\begin{document}

\title{Bent \& Broken Bicycles: Leveraging synthetic data for damaged object re-identification}

\author{Luca Piano, Filippo Gabriele Pratticò, Alessandro Sebastian Russo, Lorenzo Lanari, Lia Morra, \\ Fabrizio Lamberti  \\
Department of Control and Computer Engineering, Politecnico di Torino, Torino, Italy \\
{\tt\small \{luca.piano,filippogabriele.prattico,alessandrosebastian.russo\}@polito.it,
}\\
{\tt\small 	lorenzo.lanari@studenti.polito.it,\{lia.morra,fabrizio.lamberti\}@polito.it
}}

\maketitle
\thispagestyle{empty}

\begin{abstract}
   Instance-level object re-identification is a fundamental computer vision task,  with applications from image retrieval to intelligent monitoring and fraud detection. In this work, we propose the novel task of damaged object re-identification, which aims at distinguishing changes in visual appearance due to deformations or missing parts from subtle intra-class variations. To explore this task, we leverage the power of computer-generated imagery to create, in a semi-automatic fashion, high-quality synthetic images of the same bike before and after a damage occurs. The resulting dataset, Bent \& Broken Bicycles (BBBicycles), contains 39,200 images and 2,800 unique bike instances spanning 20 different bike models. As a baseline for this task, we propose TransReI3D, a multi-task, transformer-based deep network unifying damage detection (framed as a multi-label classification task) with object re-identification. The BBBicycles dataset is available at \url{https://tinyurl.com/37tepf7m}
   
    \textbf{keywords} instance-level retrieval; re-identification; synthetic data; damage detection; transformers
\end{abstract}

\section{Introduction}
Deep learning has fueled unprecedented advances in tasks such as person re-identification (ReID) \cite{fabbri2021motsynth,zhang2021hat,liao2021transmatcher,sharma2021person,cucchiara2022fine}, vehicle ReID \cite{khan2019survey,he2021transreid} and instance-level object retrieval \cite{zheng2017sift,amato2020large,chen2021deep,Wang2015,tan2021instancelevel}. The availability of suitable datasets for training and testing ReID systems is a key ingredient to this success.
Existing ReID benchmarks, typically focusing on persons \cite{wei2018person,zheng2015scalable,li2014deepreid} and vehicles \cite{liu2016deep,liu2017provid}, are limited in size and variety. Even when they include a large number of IDs \cite{liu2016deep,wei2018person}, they generally cover a limited geographical area (e.g., a town or campus circuit) and time window (e.g., a few hours or days). For this reason, the community has recognized the potential of synthetic data for tasks such as person detection, tracking, and ReID \cite{fabbri2021motsynth,amato2020large}. In addition to the sheer volume of generated data, synthetic generation can increase its variety in terms of background, illumination, weather, pose, etc., so that deep neural networks (DNNs) can incorporate all the invariances needed to generalize in real-world conditions. 

In the spirit of pursuing even more robust object ReID, we wish to investigate whether it is possible to make DNNs invariant not only to changes in the environment, but also to changes in the object visual appearance, such as those that could occur due to aging, degradation, damages, or removable/interchangeable parts.  Long-term ReID requires the ability to distinguish stable properties over time to account, e.g., for changes in person clothing  \cite{shu2021semantic,huang2021clothing} or seasonal changes in places \cite{masone2021survey}. Here, we propose the novel task of \textbf{damaged object re-identification}, which aims to identify the same object in multiple images even in the presence of breaks, deformations, and missing parts. Besides the theoretical interest, robust object ReID is motivated by practical applications like, e.g., fraud detection and smart contracts in the insurance domain \cite{morra2019benchmarking}.

As a benchmark for this task, we propose to focus on the study of bicycles, which are characterized by challenging intra-class variations and at the same time allow for a wide range of realistic deformations. Unlike landmarks that have unique and distinctive features, bike instances must be separated based on subtle cues (e.g., color, texture, or stickers). Deformations are inherently different from occlusions, since object parts are visible but with changes in shape (deformation) or texture (e.g., due to mud, dirt, or rust). Therefore, the insights collected from \DATASET{}
could be useful for other ReID tasks (e.g., vehicle, person), with similar challenges for long-term ReID.  Since acquiring real images of the same bicycle before and after deformation would be prohibitively challenging, we took advantage of computer graphics to generate the Bent \& Broken Bicycles (\DATASET{}) dataset, which we release as the first dataset for training and testing DNNs for damaged object ReID. 

Our \textbf{contributions} can be summarized as follows:
\begin{itemize}
\item We design a semi-automatic computer graphics pipeline to simulate different types of damage, breaks, missing parts, and material deterioration. Extensive domain randomization is further employed to train deep networks robust to variations in bicycle pose, background, etc. \cite{tobin2017domain,tang2019pamtri}. 
\item We release the \DATASET{} dataset containing 39,000 annotated images. \DATASET{} allows DNNs to (learn to) differentiate subtle intra-class variations (including different setups of the same bike model) from deformations occurring due to incidents, or aging.
\item We propose \NETWORK\ (Transformer-based object Re-IDentification \& Damage Detection), a novel transformer-based multitask DNN for joint damage detection (DD) and ReID. 
\end{itemize}

\section{Related work}
\subsection{Transformer-based re-identification}

Object ReID is the task of identifying the same object across multiple images, regardless of its pose, illumination, or context. It has many important applications such as intelligent monitoring \cite{khan2019survey,zapletal2016vehicle}, multi-object tracking and robotics \cite{morra2019benchmarking,li2018anti}, fraud detection \cite{li2018deep}, etc. The reader is referred to many comprehensive surveys for an introduction to this vast body of literature \cite{zheng2017sift,khan2019survey,chen2021deep}. In recent years, the Vision Transformer (ViT) architecture \cite{dosovitskiy2021image} has sparked a new wave of transformer-based architectures for many computer vision tasks \cite{khan2021transformers}. 
Transformer-based ReID solutions can be broadly categorized in \textit{ hybrid transformer-CNN} \cite{henkel2021efficient,zhang2021hat,liao2021transmatcher} and \textit{pure ViT-based} architectures \cite{he2021transreid,tan2021instancelevel}. 

Hybrid architectures combine CNNs as a feature extractor with a transformer-based module that tackles the matching and metric learning problem \cite{liao2021transmatcher,tan2021instancelevel,zhang2021hat,henkel2021efficient}. This approach leverages, on the one hand, CNNs hard inductive biases (e.g., translation equivariance) to work effectively on small- to medium-scale datasets. On the other hand, transformers enable cross-attention mechanisms between pairs of query and gallery images \cite{liao2021transmatcher,tan2021instancelevel}. For instance, the Reranking Transformer \cite{tan2021instancelevel} concatenates image patches from both the query and gallery images in a single sequence, which is then fed to a final classifier predicting the probability of two images representing the same object. 

More recently, a variety of pure transformer-based approaches have achieved state-of-the-art results in several ReID tasks \cite{shen2021git,zhu2021aaformer,sharma2021person,he2021transreid}. Compared to CNNs, transformers are better suited to handling long-range dependencies and avoid the use of downsampling operators (e.g., pooling and strided convolutions) that may obscure important visual details \cite{he2021transreid}.  The available architectures are typically based on a ViT backbone, pre-trained on very large-scale datasets such as ImageNet21K, and modified to extract both local and global features \cite{zhu2021aaformer,he2021transreid,shen2021git}.

\subsection{Synthetic data in deep learning}

The use of synthetic data is becoming increasingly popular for training machine and deep learning models. Although it is being experimented in multiple domains like, e.g., bioinformatics \cite{Schneider2016}, natural language processing \cite{Wang2015}, etc., this approach is indeed expected to bring the largest benefits to the field of computer vision. 
Synthetic data generation is not only an effective approach to scale data generation and annotation, it can also be used to evaluate the robustness of an algorithm under controlled conditions or to alleviate data privacy issues \cite{Zhang2016,cucchiara2022fine}.

A recent survey categorized hundreds of synthetic datasets and the use cases they have been devised for \cite{Nikolenko2021}. %
Initially used to address low-level computer vision tasks such as optical flow \cite{Mayer2015}, synthetic datasets are increasingly used to generate training datasets for high-level tasks such as, e.g., object recognition and detection \cite{Peng2015}, pose estimation \cite{Tremblay2018}, segmentation \cite{McCormac2017}, human action recognition \cite{Souza2017} and pedestrian tracking and ReID \cite{Wang2019,fabbri2021motsynth}.  Works in this field typically build onto well-known repositories, including millions of virtual models with known categories or properties, which can be programmatically manipulated to automate both data generation and its labelling \cite{Chang2015,Khodabandeh2018}. 

Popular approaches for collecting synthetic data also include the use of video games \cite{Richter2016,Courbon2010,Solovev2018,Perot2017,Morra2020}, or fusing real and virtual data via compositing techniques and placing, e.g., virtual models onto real background images \cite{Dvornik2018}.

One of the main challenges associated with synthetic data is the domain shift between real and synthetic images, which can be tackled through transfer learning or domain adaptation \cite{kar2019metasim,shrivastava2017learning,zheng2018t2net}. 
\textit{Domain randomization} is a technique used to enhance the variability of synthetic data and has been shown to substantially increase performance in the real world \cite{tobin2017domain}.
With ever increasing CGI fidelity, the synthetic-to-real domain gap is progressively reducing. Recent exciting results showed that training DNNs on very large and diverse synthetic datasets can outperform using public real datasets on tasks such as pedestrian tracking and ReID, even without fine-tuning on real data \cite{fabbri2021motsynth}.

\section{ Dataset}
\label{sec:dataset}
This section describes the semiautomatic CGI pipeline designed to generate the \DATASET\ dataset, together with its main properties and distribution. 
\begin{figure*}
\begin{center}
 \includegraphics[width=0.7\textwidth]{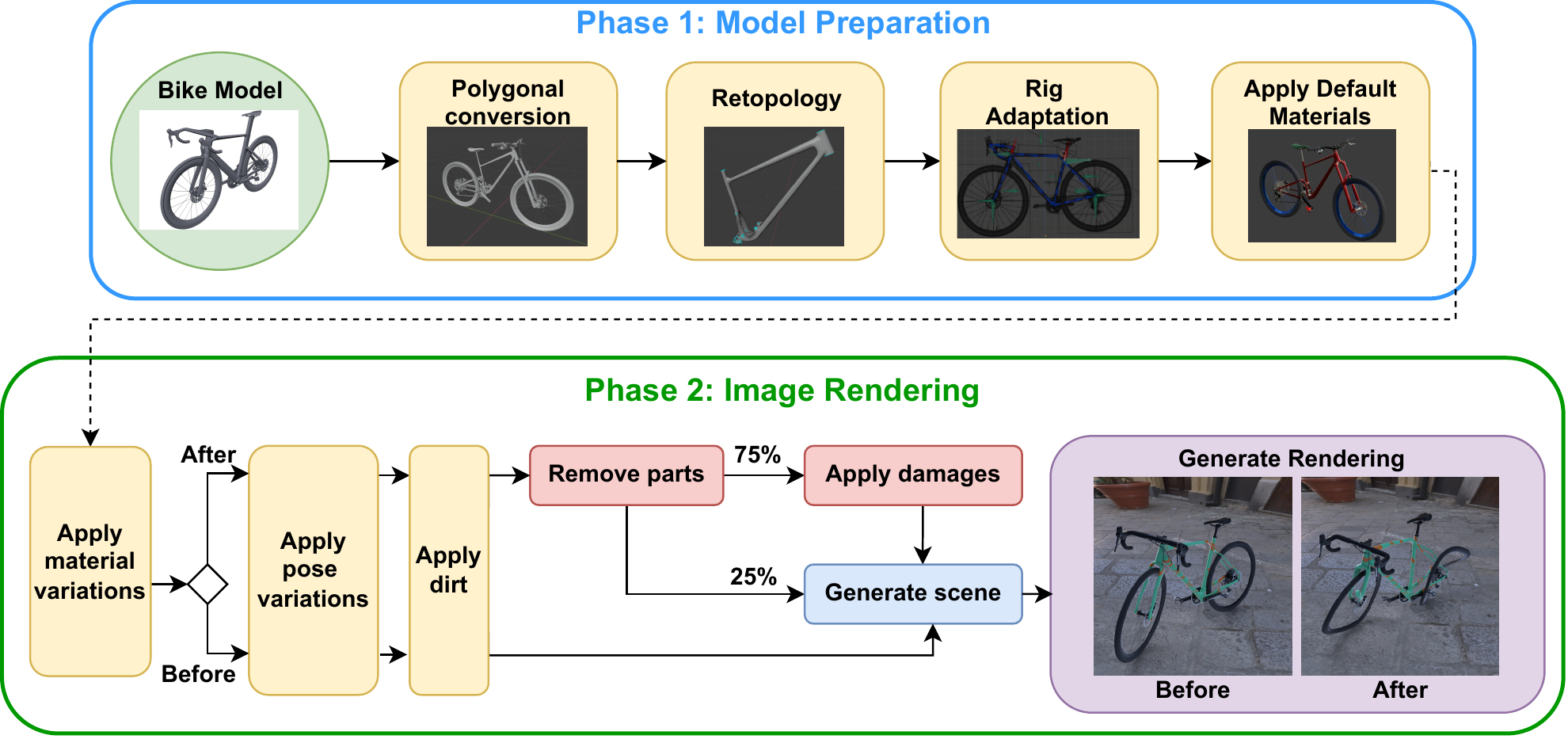}
\end{center}
   \caption{Flowchart illustrating the CGI pipeline. The 3D model is manually prepared (Phase 1) so that it can be easily manipulated by a semi-automatic image rendering script (Phase 2). The script first selects the material textures and colors, thus obtaining a new bike instance (ID). For each ID, multiple images (``before'' and ``after'' the damage) are generated, simulating damages (missing parts, bent and broken frames, etc.) with varying probabilities. Finally, the scene is generated by placing the bike onto a random background.}
\label{fig:pipeline}
\end{figure*}

\subsection{CGI Pipeline}
\label{ssec:pipeline}

The CGI pipeline, depicted in \autoref{fig:pipeline}, consists of two main phases. The first phase is model preparation, which is mostly manual and performed once for each bike model. In the second phase, a semi-automatic script generates a set of rendered images, depicting multiple views and variations of a given input bike, along with labeling and segmentation information. We sought to create a pipeline that could be applied to generate new datasets with limited human effort and hardware resources. Following this philosophy, we sacrificed some degree of photorealism in favor of reduced rendering time and increased variability.
Damages and deformations were implemented based on the classical CGI technique of 3D polygonal meshes armature deformation. 

This approach was preferred to, e.g., physics simulations, since it drastically reduces the overall rendering time while maintaining control over the output features desired in terms of missing parts, type of damage, etc. 
The whole pipeline was implemented in Blender v2.93 \cite{Blender} 
and the automatic procedure was scripted in Python as a custom add-on.

\paragraph{Model preparation.}
The input can be either a 3D parametric model (e.g., CAD file) or a polygonal mesh. In the former case, a polygonal conversion is first required to  generate a polygonal model. 
To ensure visually plausible deformations, a retopology operation has to be performed in order to obtain \emph{quad-flow} based topologies with proper vertex density in the parts that will later be subject to deformation.
Afterwards, the model is rigged and skinned (i.e., each vertex is associated with a deformation tool of the rig). To make the model easily controlled, we defined a \emph{template rig} that needs to be adapted to the given bike model. 
The template rig is made up 
of an ``armature'', ``lattices'', and ``rail guides'' (examples are shown in Appendix A). 

More in detail, the template \textbf{Armature} includes three groups/layers of ``bones''. The red bones are linked to the seat and handlebar meshes (rigid-body movement). The green bones, placed in the salient parts/joints, are used as inverse kinematic controls (targets and poles) by the blue chains. The latter are the so-called deformation bones; only this group was modified by adding/removing bones, if required by the peculiarities of the bike model. These deformation chains are the ones used for the bike frame mesh skinning, whereas other parts (e.g., seat, handlebars, and wheels) are parented (bone relatively) to the dedicated bones of the other two groups. A set of predefined deformations were devised in the form of a pose library to both change the poses of the movable bike components and introduce damages while rendering the images. 

The {\bf Lattice} 
is a three-dimensional non-renderable grid of vertices, a.k.a. deformation cage. Lattices are a convenient way of proportionally deforming a dense mesh with fewer control points since, by deforming the cage, the deformation will be transferred to the associated mesh. The lattices were used to damage the wheels. A set of deformations was devised also in this case in the form of a shape key (a.k.a. blend shape) library. 
Additionally, {\bf Rail guides} were 
used to break the bike frame exploiting a boolean mesh operation on a plane that takes the guides as reference.

\paragraph{Domain randomization and image rendering.}

After the 3D model is arranged as described, it is possible to automatically render a variety of different pictures, as described in the following. First, the 3D model is configured by randomly selecting a set of materials (texture, color, and decals) from the material library.  A physically based rendering (PBR) material library was defined, from which to pick a suitable material, among several possible choices, for each bike part.  \textit{A given combination of 3D model and materials corresponds to a single bicycle instance and is therefore assigned a unique ID}. Second, for each ID, multiple images are generated, ``before'' or  ``after'' a damage occurs,  by applying the following transformations: i) changing the pose of mobile parts (seat, handlebar, pedals and wheels); ii) (optional) applying mud or rust; iii) (optional) damage simulation; iv) point of view selection; and v) background and lighting selection. All deformations are applied randomly with predetermined probabilities and/or ranges. Possible damages include removal of one or more parts of the bike (seat, pedals, handlebar, and wheels), bent frame, broken frame, and wheel deformation.

Finally, the rendered bike must be placed onto a suitable background, adjusting for the specific lighting conditions. 

The approach considered in the pipeline takes advantage of the LilyScraper \cite{Lily}, a Blender add-on to use a High Dynamic Range Imaging (HDRI) map as background and light source, in combination with a shadow-catcher plane. The setup of the environment and the lighting was performed once for all models.

\subsection{\DATASET\ characteristics}
\label{ssec:datadistrib}

\paragraph{Dataset distribution.}
The final dataset contains a total of 39,200 images from 2,800 unique IDs (20 models, 140 IDs each). 
20 models retrieved from dedicated marketplaces were prepared, including 6 MTBs, 1 Enduro, 6 Road bikes, 1 Circuit, 1 Gravel and 5 Cruiser (following the categorization introduced in \cite{regenwetter2022biked}). For the textures, we collected five patterns of various styles. Both the base and pattern colors were randomly chosen from a pool of 50 colors. Additionally, 10 different decals containing logos from famous bike brands such as \emph{Bianchi} and \emph{Cannondale} were randomly applied. The background was selected from a pool of 11 different 360° HDRIs, varying bike positioning and illumination by rotating the camera.  

For each bike ID, up to 14 renderings were generated, evenly divided in ``before'' and ``after'' images as shown in the flowchart (\autoref{fig:pipeline}). For ``before'' images, only dirt or rust was applied with 20\% probability. For ``after'' images, dirt/rust was applied with 50\% probability, damages to the frame were applied with 75\% probability (25\% were bent, 25\% were broken and 25\% were both bent and broken), and finally each removable part (seat, pedals, handlebar, and wheels) was removed (50\% probability) or deformed (50\% probability). Thus, some of the ``after'' images are not damaged.  
Labels for the ReID task were automatically generated based on the bike unique ID assigned by the pipeline.

\paragraph{Training, validation, and stress test set.}
The dataset was split into a training, validation, and test set at the level of bike ID and model to test DNNs' ability to generalize both across IDs and across models. The validation set includes both models seen and unseen during training, whereas the (stress) test set includes only models that were never seen in either the training or validation set, to ensure that it is sufficiently challenging and representative of real operating conditions. Specifically, the training set contains 25,676 images (1,834 IDs, 14 models), the validation set contains 1,128 images (564 IDs, 12 models), and the stress test contains 840 images (420 IDs, 3 models).

\paragraph{Real dataset.}
A separate dataset of real photos of damaged and undamaged bikes was also collected to test the ability of \NETWORK\ to generalize to the real domain. 
We combined a subset of the publicly available DelftBikes dataset
\cite{kayhan2021hallucination} with images collected by web scraping from popular search engines and e-commerce sites. The images were manually labeled following the same criteria as those used for the synthetic dataset. 
A total of  6,292 images were collected, of which 106 presented a Bent (64) or Broken (52) frame. 
The dataset was split into train, validation and test with a 7:1.5:1.5 split, stratified by damage type.  

\section{Methodology}
\label{sec:Architecture}

\begin{figure*}[tbh]
\begin{center}
 \includegraphics[width=0.75\textwidth]{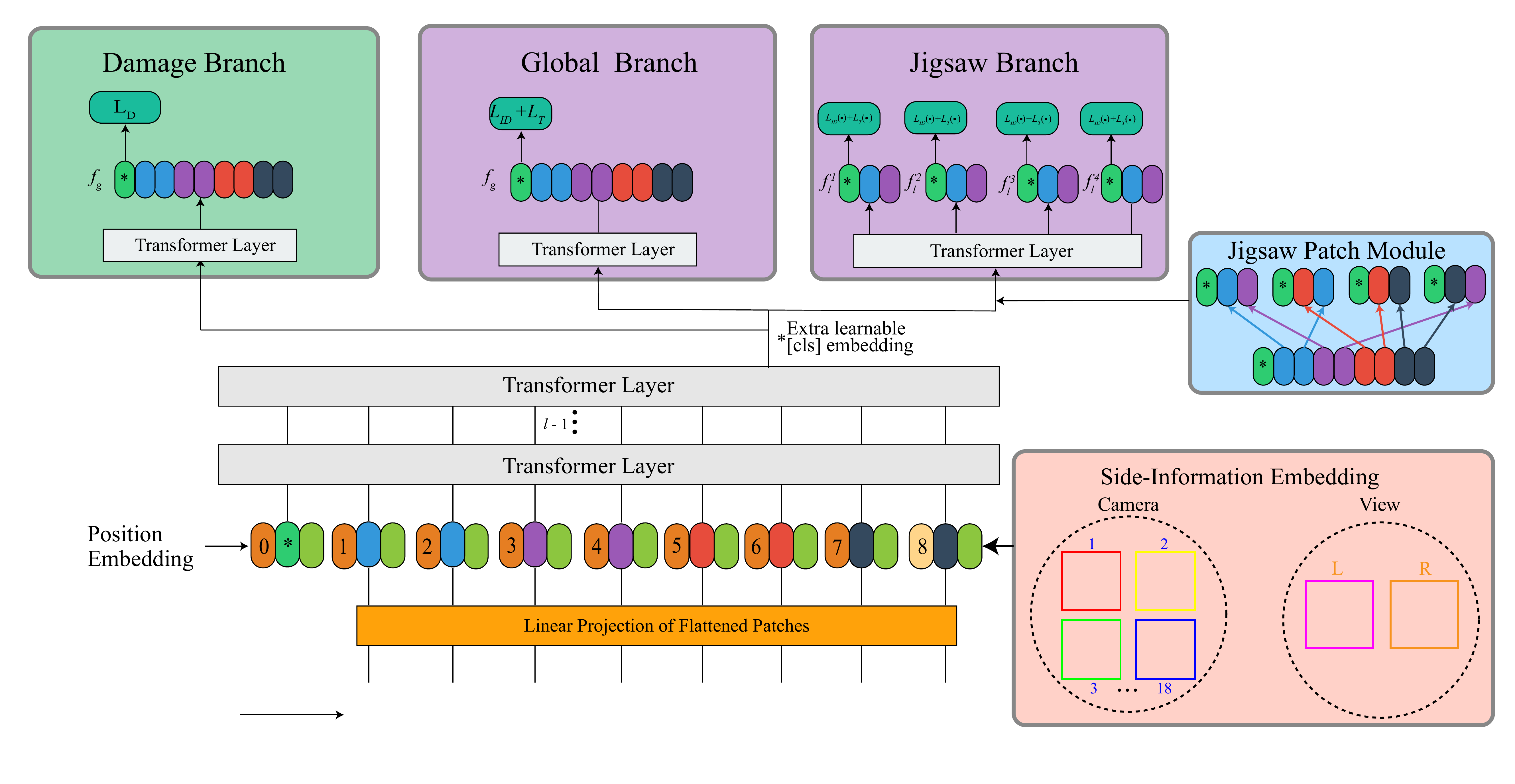}
\end{center}
   \caption{\NETWORK\ architecture. Embeddings are enriched with  position and camera information (side information embedding). A learnable \CLS\ token is prepended to the embeddings which are input to a shared backbone. Task-specific branches (DD branch, Global ReID branch and Jigsaw Branch with JPM) include a separate transformer layer to adapt global features to each task. The Jigsaw Module, $\mathcal{L}_{I D}$, and $\mathcal{L}_{T}$ are described in \cite{he2021transreid}. 
   }
\label{fig:TDReID_scheme}
\end{figure*}

\paragraph{Problem setting}

We assume that the training set $D$ consists of $N$ sequences of synthetic images $D ={\{(x^1_i, ... , x^M_i)\}}_{i=1}^N$, where all images $x^j_i$ in a sequence are associated with the same ID $i$ and represent the same bike instance. We additionally assume that each image is associated with a set of binary attributes, each representing the presence of a specific kind of damage ( $ a^j_i \in \mathcal{A} = \{BD, BK, P_n\}$); $P_n$ indicates whether the $n^{th}$ part is present or missing. Given $D$, our aim is to learn an embedding space $ x^j_i \in  \mathbb{R}^{h\times w \times ch}  \mapsto e^j_i \in \mathbb{R}^{m}  $ such that all images associated with a given ID $i$ are closer in the embedding space than other IDs, regardless of the attributes $a^j_i$. We further define the DD task as predicting the values of $a^j_i$ (multi-label binary classification). At inference time, a query image is compared against the gallery, and the correct ID must be retrieved on the basis of the embedding distance. We assume that the damaged bikes are the queries, inspired by applications in the insurance domain (fraud detection).

\paragraph{\NETWORK{} architecture}

The \NETWORK{} architecture for joint DD and ReID, shown in \autoref{fig:TDReID_scheme}, builds on the TransReID \cite{he2021transreid} architecture, which achieved state-of-the-art performance among ViT-based models for vehicle ReID, and enriches it with an additional multi-label DD branch. 

The TransReID architecture \cite{he2021transreid} builds on the ViT architecture \cite{dosovitskiy2021image}, but includes additional components to capture more robust and fine-grained features. Specifically, the Side Information Embedding (SIE) module encodes non-visual information such as camera or viewpoint, and is input to a transformer encoder together with learnable patch and position embeddings. The global ReID branch and the Jigsaw branch then jointly learn the ReID task, encoding global ($f_g$) and local ($f_l$) features, respectively. The Jigsaw branch is based on the Jigsaw Patch Module (JPM), which shuffles all patches and regroups them into several groups, all of which are input to a shared transformer layer to learn local features $f_l$, as detailed in \cite{he2021transreid}. 

\paragraph{Damage branch and multi-task learning.}

Multi-task learning is implemented using one shared transformer backbone and an additional separate transformer layer for each task \cite{peng2020empirical,wolfe2021exceeding}.
The DD branch is a multi-label classifier with seven output heads: two for Bent and Broken frame labels, and five for missing parts (front wheel, rear wheel, seat, handlebar or pedals). Each output head takes as input the \CLS\ token and passes it through a batch normalization (BN) layer followed by a fully connected (FC) layer. 

\NETWORK\ combines two tasks, one executed on image pairs (ReID) and one executed on individual images (DD). In addition, the ReID task is not defined for real images. For this reason, a multi-task diversion mechanism was implemented which selects the tasks that need to be executed upon the extracted features of each training batch. Hence, synthetic images are forwarded to all branches, whereas real images are directed to the DD branch only.

\paragraph{Loss computation.}
The loss combines the ReID loss, including global and local features, with the DD loss: 

   \begin{equation}    
    \begin{split}
    \mathcal{L}=\alpha\mathcal{L}_{I D}\left(f_{g}\right)+\beta\mathcal{L}_{T}\left(f_{g}\right)+\gamma\mathcal{L_{D}}\left(f_{g}^a, f_{g}^p, f_{g}^n\right) \\
    + \frac{1}{k} \sum_{j=1}^{k}\left(\mathcal{L}_{I D}\left(f_{l}^{j}\right)+\mathcal{L}_{T}\left(f_{l}^{j}\right)\right)
    \end{split}
    \end{equation}

where $\mathcal{L}_{T}$ and $\mathcal{L}_{ID}$ are the triplet loss and the ID cross-entropy loss (which treats each ID as a separate class, as defined in \cite{he2021transreid}),

${\mathcal{L}_{D}}$ is the DD loss, and $k$ (= 4) is the number of classification heads of the JPM branch. All loss components are calculated on the \CLS{} token ($f_g$: global branch, $f_l$: Jigsaw branch). To compute $\mathcal{L}_{T}$, triplets are online sampled from each batch with hard negative and positive mining. 
${\mathcal{L}_{D}}$ is a weighted binary cross-entropy loss:   
    \begin{equation}    
    \begin{split}
    \mathcal{L}_{D}=\lambda\mathcal{L}_{BD}\left(\cdot\right)+\mu\mathcal{L}_{BK}\left(\cdot\right)   +\nu\frac{1}{n} \sum_{j=1}^{n}\left(\mathcal{L}_{P_{n}}\left(\cdot\right)\right)
    \end{split}
    \end{equation}
where $\mathcal{L}_{BD}$ and $\mathcal{L}_{BK}$ refer to the Bent and Broken frame labels losses, and $\mathcal{L}_{P_n}$ to the $n=5$ specific missing parts losses. In the case of real images, for the sake of simplicity we consider only $\mathcal{L}_{BD}$ and $\mathcal{L}_{BK}$.

\paragraph{Domain adaptation.}
In the baseline, \NETWORK{} is trained on \DATASET{} and tested on the real data set, without adaptation or fine-tuning. We further explored different domain adaptation strategies. For \textit{supervised domain adaptation}, we simply leveraged the multi-task training strategy to train the model on real and synthetic data. For \textit{unsupervised domain adaption}, we experimented with the well-known domain adversarial technique DANN \cite{ganin2016domain} and with partial domain adaptation PADA \cite{cao2018partial}. Experiments with PADA were motivated by the observation that \DATASET\ includes a wider range of bike models and setups compared to the real dataset, and therefore forcing the feature distributions to align could lead to negative transfer. PADA assumes that the target domain contains different labels than the source, whereas in our setting DD labels are the same (additionally, all labels are binary given the multi-label setting). Therefore, we introduced the auxiliary task of bike model classification  (model information is available for synthetic images); PADA exploits these predictions to enhance the contribution of (samples of) bike models that are present both in the synthetic and real datasets. Further details are available in Appendix B. 

\section{Experimental settings}
\label{sec:Experiments}

\paragraph{\NETWORK{} Training and hyper-parameter settings.}
All images were resized to $256 \times 256$, and normalized with the mean and standard deviation calculated on the synthetic training set. Data augmentation was performed with random color- and texture-preserving transformations (horizontal flip, crop, blurring, and gaussian noise). Each image was split into overlapping  $16 \times 16$ patches, with patch stride set to $12 \times 12$. Batches containing either real or synthetic images were alternated, and the real dataset was iterated twice per epoch to counterbalance the smaller size.

For all experiments, the model backbone was pre-trained on ImageNet \cite{deng2009imagenet}, and the remaining weights were initialized by Kaiming normal initialization \cite{he2015delving}. All models were trained for 20 epochs. The SGD optimizer was used with batch size set to 32, momentum to 0.9 and weight decay to 1e-4. The cosine learning rate scheduler was used (initial learning rate 0.01, linear warmup for 5 epochs). 

Regarding the loss, we set $\alpha$, $\beta$ and $\gamma$ to 1, whereas for $\mathcal{L}_{D}$, we set  $\lambda$, $\mu$, $\nu$ to 0.25, 0.25 and 0.5, respectively.

\paragraph{Other baselines.}
\NETWORK{} was compared against the Reranking Transformers (RRT) Global retrieval baseline \cite{tan2021instancelevel}. RRT was trained on \DATASET for 50 epochs. The training setting is the same as the default one used in the original code, with learning rate of 1e-3, SGD optimizer with 0.9 momentum, batch size 128, weight decay of 4e-4, MultiStep learning rate scheduler with a 0.1 decay at epochs 30 and 40, contrastive loss and ResNet-50 backbone. However, since RRT does not perform damage detection, it was evaluated only on the ReID task.

\paragraph{Evaluation protocol.}
Performance on the ReID task was measured using common metrics for vehicle and object ReID, i.e., mean Average Precision (mAP) and Cumulative Matching Characteristics (CMC) \cite{he2021transreid}. CMC-$K$, with $K=\{1,5,10\}$, represents the average probability of observing the correct identity within the top-$K$ ranked results. Since the gallery contains one instance per bike ID, it is equivalent to Recall@K. For each pair of images in the validation and stress test, we set the ``after'' image as Query and the ``before'' image as Gallery. All images from other IDs (including those derived from the same 3D bike model) were used as distractors. 

For the DD task, performance was measured using the Area under the Receiver Operating Characteristic Curve (AUROC), macro-averaged across all labels. For the sake of conciseness, we report only results for Bent and Broken labels, since damages to the frame are more challenging to detect than missing parts. All performance metrics were averaged over three runs. 
\section{Results}

\begin{table*}[tbh!]
    \begin{center}
        \begin{tabular}{c|cccccc|}
        \cline{2-7}
        & \multicolumn{6}{c|}{\cellcolor[HTML]{EFEFEF}\textbf{Validation}}  \\ 
        \cline{2-7} 
        
        & \multicolumn{2}{c|}{\textbf{Damage Detection}}             & \multicolumn{4}{c|}{\textbf{Re-identification (Synthetic)}} \\ 
        \cline{2-7} 
         & \multicolumn{1}{c|}{\textbf{Real AUC}} &
          \multicolumn{1}{c|}{\textbf{Synthetic AUC}} &
           \multicolumn{1}{c|}{\textbf{mAP}} &
          \multicolumn{1}{c|}{\textbf{CMC-1}} &
          \multicolumn{1}{c|}{\textbf{CMC-5}} &
          \textbf{CMC-10} \\ 
         \hline
                 \multicolumn{1}{|c|}{BL}         & 93.4 ± 1.5 & \multicolumn{1}{c|}{92.1 ± 0.5}        &  \textbf{85.3 ± 0.2}   & \textbf{79.8 ± 0.5}  & 91.9 ± 1.1          & 96.3 ± 0.5          \\
        \multicolumn{1}{|c|}{BL + Real\textsuperscript{$\dagger$}}
        & \textbf{97.3 ± 2.2} & \multicolumn{1}{c|}{91.4 ± 0.2}       & \textbf{85.3 ± 0.2}    & 79.4 ± 0.1           & \textbf{92.9 ± 0.4}          & \textbf{96.6 ± 0.4} \\
        \multicolumn{1}{|c|}{RRT (Global)}         & - & \multicolumn{1}{c|}{-}        &  80.5 ± 1   & 74.1 ± 1.6 &  88.3  ± 1.1         & 93.4 ± 1.2          \\ \hline
        \multicolumn{1}{|c|}{BG Places365 + Real \textsuperscript{$\dagger$}}     & 96.3 ± 1.9 & \multicolumn{1}{c|}{90.4 ± 0.2}  & 85 ± 0.1       & 79.0 ± 0.4           & 92.8 ± 0.3          & 96.3 ± 0.2          \\
        \multicolumn{1}{|c|}{BG Uniform + Real \textsuperscript{$\dagger$}}       & 95.2 ± 3.4 & \multicolumn{1}{c|}{87.4 ± 1.5}  &  48.5 ± 3.4       & 39.2 ± 1.9           & 59.4 ± 5.6          & 66.0 ± 5.7          \\ \hline
        \multicolumn{1}{|c|}{ReID (single task)\textsuperscript{$\dagger$}}        & -          & \multicolumn{1}{c|}{-}         & 83.3 ± 1.2           & 77.0 ± 1.2             & 91.2 ± 1.5          & 95.1 ± 1.4          \\ 
        \multicolumn{1}{|c|}{Damage detection (single task)\textsuperscript{$\dagger$}}      &  \textbf{97.5 ± 1.5} & \multicolumn{1}{c|}{\textbf{94.5 ± 0.5}}  & - & -                    & -                   & -                   \\ \hline
        \multicolumn{1}{|c|}{BL + DANN\textsuperscript{$\ddagger$}}        & 93.9 ± 1.1 & \multicolumn{1}{c|}{91.7 ± 0.9}       & 85.2 ± 0.2    & 79.4 ± 0.4           & 92.3 ± 1.0  & 96.4 ± 0.5             \\
        \multicolumn{1}{|c|}{BL + Real + DANN\textsuperscript{$\dagger$}} & 97.0 ± 1.8   & \multicolumn{1}{c|}{91.0 ± 0.6}       & 85.2 ± 0.5      & 78.9 ± 0.8           & 92.8 ± 0.4 & 96.4 ± 0.7          \\ \hline
        \multicolumn{1}{|c|}{BL + PADA \textsuperscript{$\ddagger$}}        & 94.4 ± 0.5 & \multicolumn{1}{c|}{90.8 ± 1.2}    & 84.8 ± 0.2       & 78.6 ± 0.4           & 92.6 ± 0.3   & 96.4 ± 0.5     \\
        \multicolumn{1}{|c|}{BL + Real + Model labels\textsuperscript{$\dagger$}} & 96.9 ± 1.9 &   \multicolumn{1}{c|}{90.7 ± 1.0}  & 84.6 ± 0.4 &   77.9 ± 0.7 &   \textbf{93.0 ± 0.4} & 96.6 ± 0.1 \\
        \multicolumn{1}{|c|}{BL + Real + PADA\textsuperscript{$\ddagger$}} & 96.2 ± 3.1 & \multicolumn{1}{c|}{90.9 ± 1.9}  & 84.7 ± 0.1  & 78.4 ± 0.2           & 92.4 ± 0.3          & \textbf{96.9 ± 0.6}          \\ \hline
                \cline{2-6}
          & \multicolumn{6}{c|}{\cellcolor[HTML]{EFEFEF}\textbf{Stress test}}                                            
        \\ \cline{2-7} \hline     
         \multicolumn{1}{|c|}{Baseline}        & - &  \multicolumn{1}{c|}{94.1 ± 0.2}         & \textbf{79.3 ± 0.2} & \textbf{72.5 ± 0.2}     & 87.4 ± 0.3     & \textbf{92.2 ± 0.1}    \\
         \multicolumn{1}{|c|}{BL + Real\textsuperscript{$\dagger$}}    & -    &  \multicolumn{1}{c|}{93.5 ± 0.23}   & 79.2 ± 0.1      & 72.1 ± 0.4              & 88.0 ± 0.1      & \textbf{92.2 ± 0.1}    \\ 
                        \multicolumn{1}{|c|}{RRT (Global)}     & -    &  \multicolumn{1}{c|}{ - }     &   76.1 ± 1.3 & 65.7 ± 2.3     & 85.4 ± 2.2     & 90.6 ± 0.9    \\
         \hline
         \multicolumn{1}{|c|}{BL + DANN\textsuperscript{$\ddagger$}}     & -      & \multicolumn{1}{c|}{93.4 ± 1.1}         & 78.7 ± 0.5  & 71.6 ± 0.5              & 87.9 ± 0.5    & 91.3 ± 0.7             \\
         \multicolumn{1}{|c|}{BL + Real + DANN\textsuperscript{$\dagger$}}     & -   & \multicolumn{1}{c|}{93.5 ± 0.3}       &  79.1 ± 0.2 & 71.7 ± 0.2              & 87.9 ± 0.2    & 92.1 ± 0.2             \\ \hline
         \multicolumn{1}{|c|}{BL + PADA\textsuperscript{$\ddagger$ } }    & -   &  \multicolumn{1}{c|}{\textbf{94.2 ± 0.4}} & 79.2 ± 0.4 & 72.3 ± 0.8    & \textbf{88.1 ± 0.1}     & \textbf{92.2 ± 0.5}    \\

         \multicolumn{1}{|c|}{BL + Real + PADA\textsuperscript{$\dagger$}}   & -  & \multicolumn{1}{c|}{92.9 ± 1}    &     78.9 ± 0.7   & 71.9 ± 0.7              & 87.8 ± 0.8    & 91.9 ± 0.1             \\ \hline
        \end{tabular}
    \end{center}
    \caption{Performance on the validation and stress set. All networks trained on synthetic data except for $\dagger$ (labeled real images available at training time) and $\ddagger$ (unlabelled real images available at training time). Best results are in bold. }
    \label{tab:results1}
\end{table*}

\paragraph{What is the DD and ReID performance of the baseline, with and without real labeled images at training time?}
The baseline was trained in two different settings: one assuming that only synthetic data is available at training time (BL), and one assuming that a small sample of labeled images is available at training time (BL+Real). 
As shown in \autoref{tab:results1}, \textit{on the DD task} BL achieves an average AUC of 92.1 ± 0.5 for synthetic images and of 93.4 ± 1.5 for real images. 
However, we postulate that there is still a domain shift between the synthetic and the real data, since performance on the DD task improved when the network was exposed to the real domain during training (AUC=97.3 ± 2.2). 

Delving deeper in the DD task, \textit{the performance varies for different damage types on the synthetic dataset}, with higher AUC on Broken (100 ± 0.0) than Bent frames (81.5 ± 2). Bent frames are more challenging to detect since some frames (e.g., Cruiser) may include both straight and curved lines, and \DATASET{} includes a range of both subtle and heavy damages. On the other hand, the visual features associated with broken frames are well defined and stable between different bike models. 

\begin{figure*}
\begin{center}
 \includegraphics[width=\textwidth]{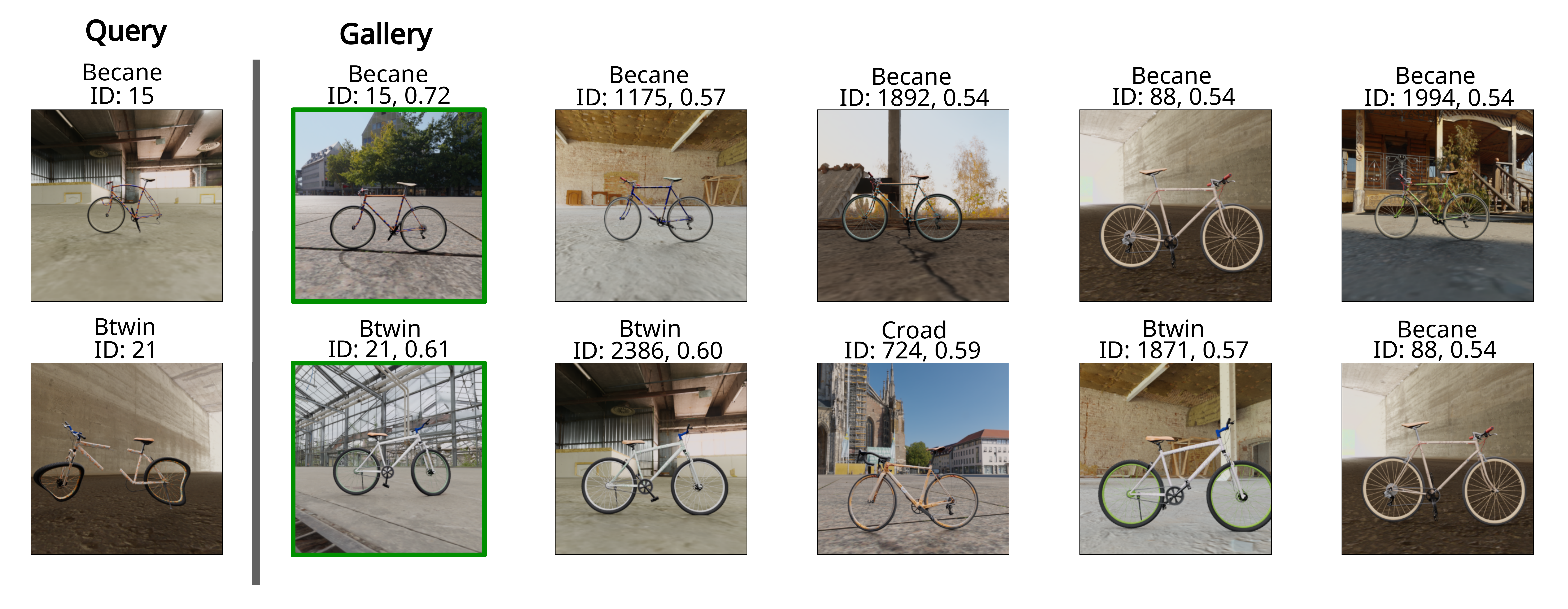}
\end{center}
\vspace{-6mm}
   \caption{Retrieval results (Top-5 images) for the BL network (ID and similarity scores). 
    The correct ID is retrieved despite the presence of missing parts (ID 15), bent  (ID 15) or broken (ID 21) frame, deformed wheels (ID 21), and rust (ID 21).}
\label{fig:retrieval}
\end{figure*}

\textit{On the ReID task,} \NETWORK{} achieved a mAP of 85.3 ± 0.2 (BL and BL+Real) and a CMC-1 of 79.8 ± 0.5 (BL) and 79.4 ± 0.1 (BL+Real), with minor variations when exposed to real data during training. \autoref{fig:retrieval} shows how \NETWORK{} is able to predict the correct ID and distinguish damage-induced variations from different setups of the same (or similar) bike models. 

We further investigated the \textit{effect of the background} on the ReID and DD performance. Specifically, we compared three choices of background: (i) HDRI images, as detailed in Section \ref{sec:dataset}; (ii) random selection from Places365 \cite{zhou2017places}, and (iii) a simple uniform background (see Appendix C for examples).  On the DD task, all transfer scenarios ($ \text{HDRI (BL)} \to \text{Real} $, $ \text{ BG Places365} \to \text{Real} $ and $ \text{BG Uniform} \to \text{Real}$) achieved similar results (\autoref{tab:results1}). HDRI slightly outperforms Places365: the latter contains a wider range of scenes, but the resulting blend is not as realistic as the proposed HDRI technique. On the ReID task, performance substantially drops when training on a uniform background, as the network does not learn to separate the bike from the background.

\paragraph{Is multi-tasking beneficial for damaged object re-identification?}
We compared \NETWORK{} against single-task ReID and DD networks -- the former reduces to the original TransReID architecture, whereas the latter becomes a ViT-based multi-label classifier. As shown in \autoref{tab:results1}, \NETWORK{} outperforms the single-task ReID architecture both in terms of mAP (85.3 ± 0.2 vs. 83.3 ± 1.2) and CMC (CMC-1 79.9 ± 0.4 vs. 77.0 ± 1.2). This is further confirmed by the performance of RTT (mAP 80.5 ± 1 vs. 85.3 ± 0.2). On the other hand, DD improves in the single-task setting on both real (97.5 ± 1.5) and synthetic (94.5 ± 0.5) images. A possible explanation is that the ReID task forces the network to take into account the entire bicycle, whereas for DD simpler, more localized visual cues are sufficient. Conversely, the ReID task can leverage the DD labels to learn visual properties invariant to the presence of damage.

\paragraph{Are feature-level domain adaptation strategies helpful to reduce the synthetic-to-real gap?}
The BL results indicate that, at least for the DD task, a certain domain shift still exists. Besides low-level differences due to CGI, we postulate that this domain shift may be attributed to different reasons: on the one hand, few examples of damaged bikes are available; on the other hand, the synthetic dataset contains more bike models (for instance, most images in the Delft Bikes dataset are minor variations of a typical city bike). As detailed in Section \ref{sec:Experiments}, we have tested two techniques, DANN and PADA, focusing on the DD task. %

When labeled real images are available during training, neither DANN (97.0 ± 1.8) nor PADA (96.2 ± 3.1) outperforms BL + Real (97.3 ± 2.2). On the other hand, if we assume that labels are not available at training time, both DANN (93.8 ± 1.1) and PADA (94.4±0.5) improved over BL (93.4 ± 1.5), but did not match the supervised setting (97.3 ± 2.2). On the ReID task, domain adaptation slightly hurts the performance in terms of CMC-1, bearing however in mind that this task is evaluated only on synthetic images. t-SNE plots of the \CLS{} token extracted from the backbone (Appendix C) show only partial overlap between the real and synthetic domains. Saliency (attention) maps generated following the approach in \cite{chefer2021generic} highlight how the network correctly focused its attention on the bike frame (and occasionally the wheels) (\autoref{fig:xai}). Different training regimes consistently yield similar visual keys (Appendix C).

\paragraph{How does the network generalize to previously unseen bike models?}

Overall, the DD task generalizes well to previously unseen models, while performance is more dependent on the specific type of damage. When training on synthetic data alone (BL), we observed an increase in performance for the DD task from 92.1 ± 0.2 to 94.1 ± 0.2 (\autoref{tab:results1}). Again, forcing the network to improve on real images lowers the performance on synthetic images for all strategies but BL + PADA (94.2 ± 0.4). However, the latter incorporates an additional bike model classification task, which may help \NETWORK{} to better generalize to previously unseen models. On the other hand, in the ReID task both \NETWORK{} and RRT struggle to generalize to completely novel bike models, with a moderate decrease in performance both in terms of mAP (79.3±0.1 vs. 85.3±0.2) and CMC-1 (72.5±0.2 vs. 79.8±0.5).

\begin{figure}[t]
    \begin{center}
    \includegraphics[width=0.6\columnwidth]{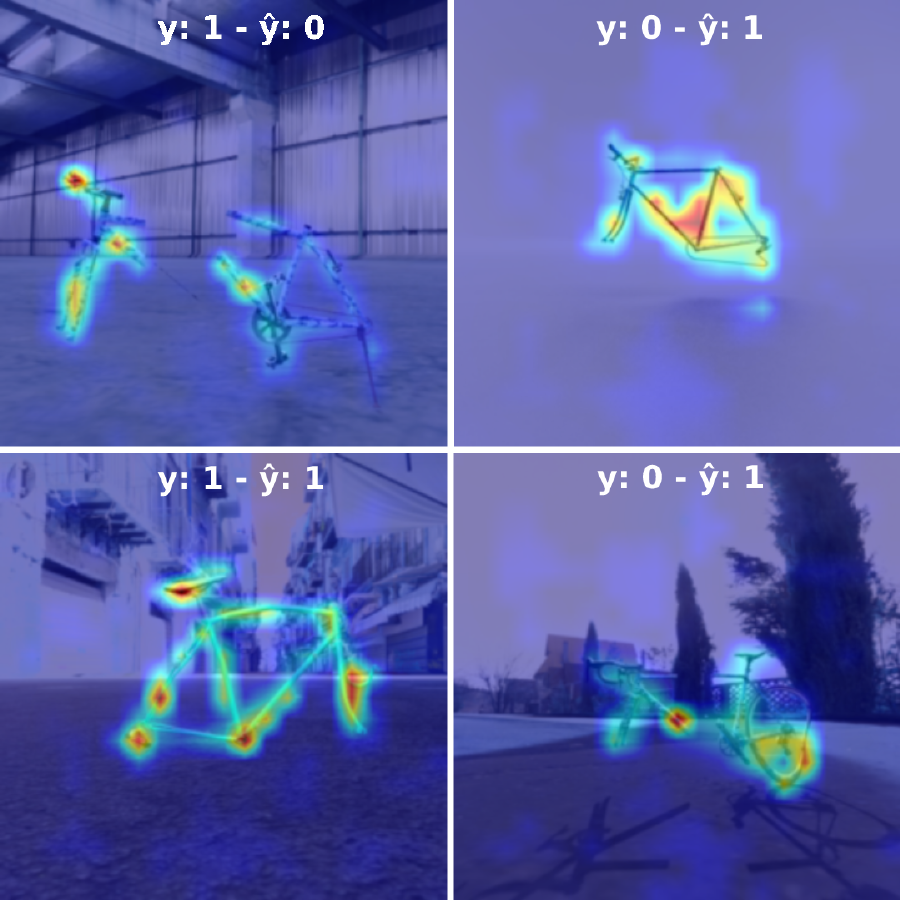}
    \caption{Attention maps of \NETWORK{} for BL + REAL + DANN, with Bent frame labels (y) and predictions ($\hat{y}$). }
    \label{fig:xai}
    \end{center}
\end{figure}

\section{Conclusions}
In this work, we introduced the novel task of damaged object re-identification. As a benchmark for this task, we introduced the synthetic \DATASET{} dataset which contains paired images of the same bike with and without damages.  As a baseline, we proposed \NETWORK{}, a multi-task tranformer-based architecture for joint DD and ReID. Experimental results showed how the DD task improves performance on the ReID task, but not viceversa.  
The main limitation of the present work is the lack of real paired images of bikes, before and after damage; for this reason, only the DD task was analyzed for real images. As collecting such a dataset would be prohibitively expensive, an option to be explored is simulation, e.g., through data augmentation or generative models. Given the novelty of the task, there is ample room for future expansion in several directions. First, concerning the ReID task, the ability to generalize to previously unseen models should be improved. Experiments should also be extended to include more traditional convolutional architectures. Second, techniques for bridging the synthetic-to-real gap could be further investigated, e.g. by looking at the few-shot and partial/universal domain adaptation literature. Third, segmentation could be leveraged to improve foreground/background differentiation. 
Finally, other tasks could be explored using the proposed pipeline and the collected 3D models in combination with rendered images, e.g., cross-modal image retrieval \cite{jing2021cross,uy2020deformation}, segmentation, and 3D part recognition \cite{yao2021discovering}.
\section*{Acknowledgements}
The authors gratefully acknowledge the financial support of Reale Mutua Assicurazioni.
    
{\small
\bibliographystyle{ieee_fullname}
\bibliography{main}
}

\appendix
\onecolumn
\section{Dataset}
\subsection{CGI Pipeline implementation details}

In this section we provide additional details on the CGI pipeline used to implement the \DATASET{} dataset, and in particular on the transformations that are randomly applied to generate ``before'' and ``after'' images.

\subsubsection{Template rig. }

\begin{figure}[h] 
    \centering
    \subfloat[]{\includegraphics[height=2.7cm]{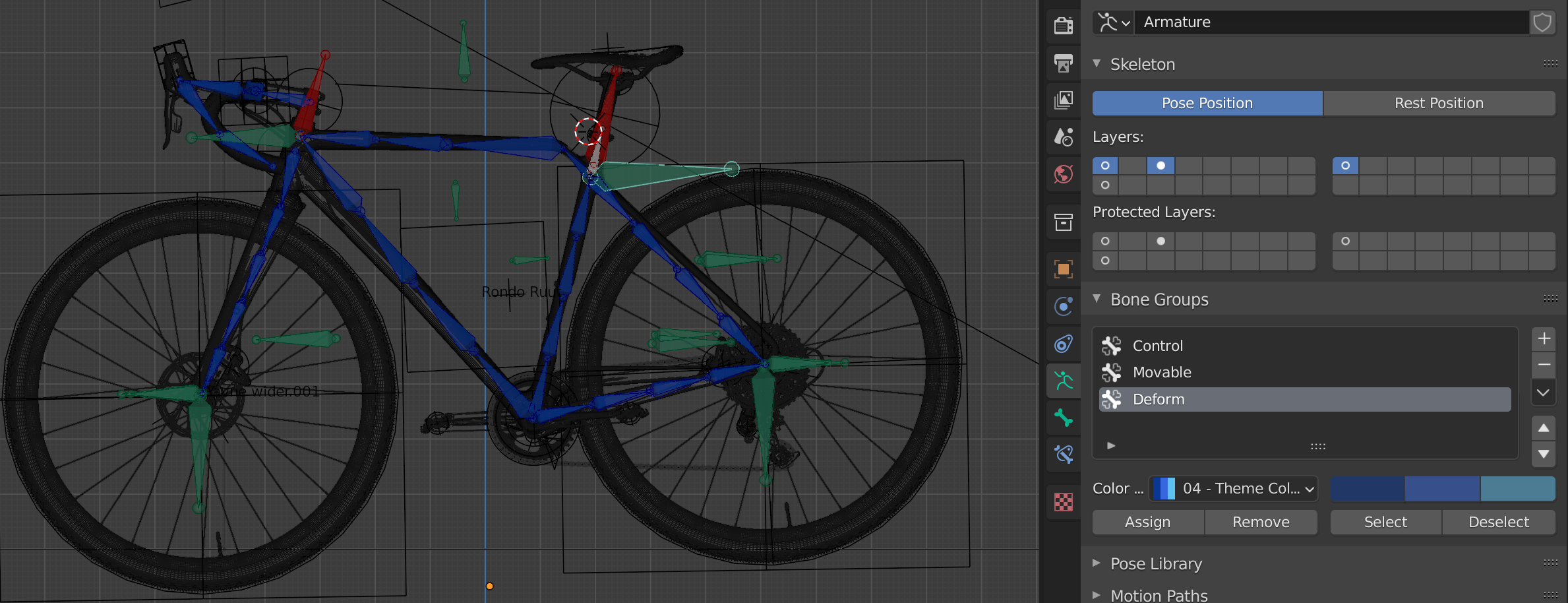}%
   \label{fig:rig_1}}
    \hfill\subfloat[]{\includegraphics[height=2.7cm]{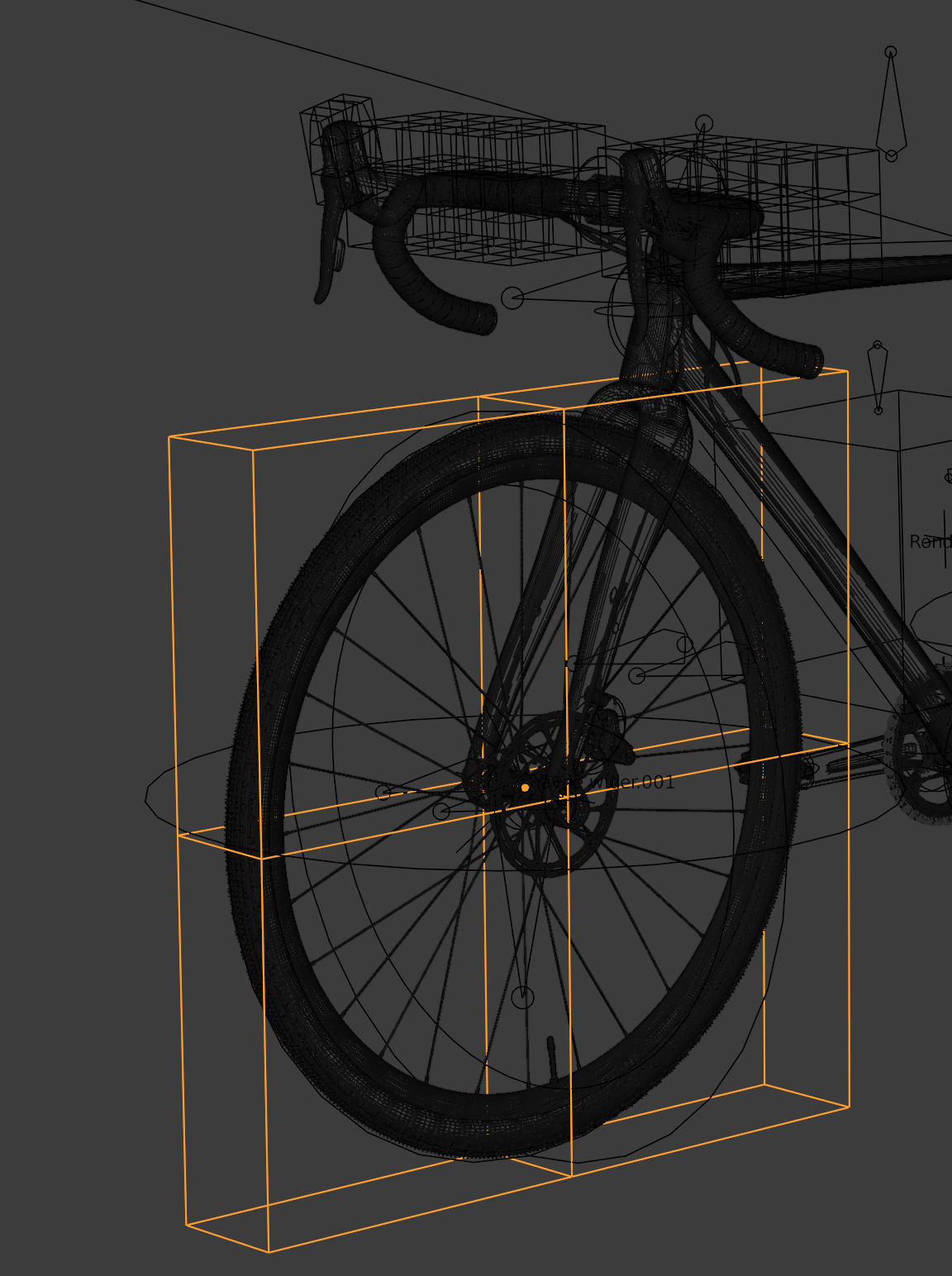}%
    \label{fig:rig_2}}
    \hfill\subfloat[]{\includegraphics[height=2.7cm]{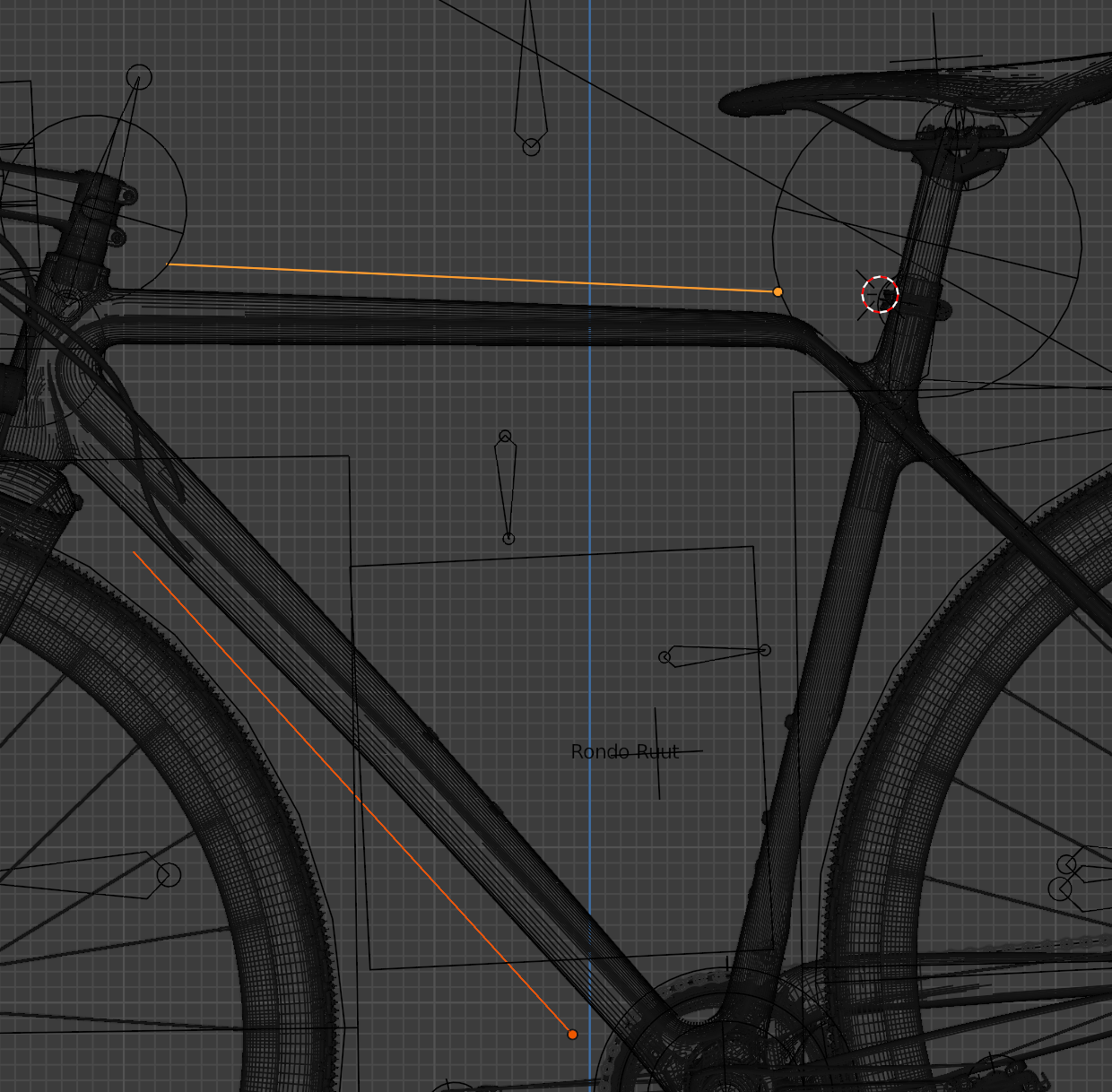}%
    \label{fig:rig_3}}
    
   \caption{Template rig adaptation and skinning: (a) armature with bone groups and layers, (b) fitting lattice to wheel meshes, and (c) rail guides positioning.
    }
    \label{fig:rig}
\end{figure}

A bike contains many movable elements which need to be positioned (to randomly change the bike pose) or deformed (to simulate damages). In order to randomly apply these transformations,  we defined a \emph{template rig} that needs to be adapted to match the given bike model. The template rig is composed of an ``armature'' (\autoref{fig:rig_1}),  ``lattices'' (\autoref{fig:rig_2}), and ``rail guides'', placed as depicted in \autoref{fig:rig_3}.

\subsubsection{Image rendering.}
\begin{figure}[tbh] 
    \centering
    
\subfloat[Intact frame]{\includegraphics[width=1.0\textwidth]{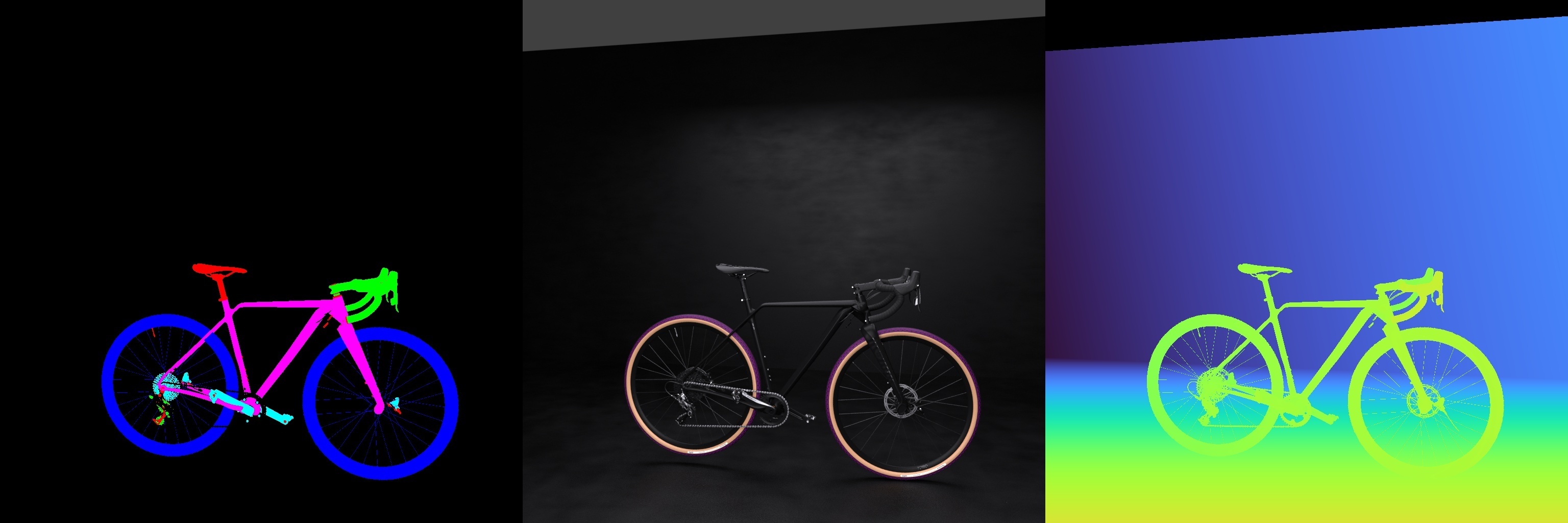}%
\label{fig:ex_3}}
\vfill\subfloat[Broken frame]{\includegraphics[width=1.0\textwidth]{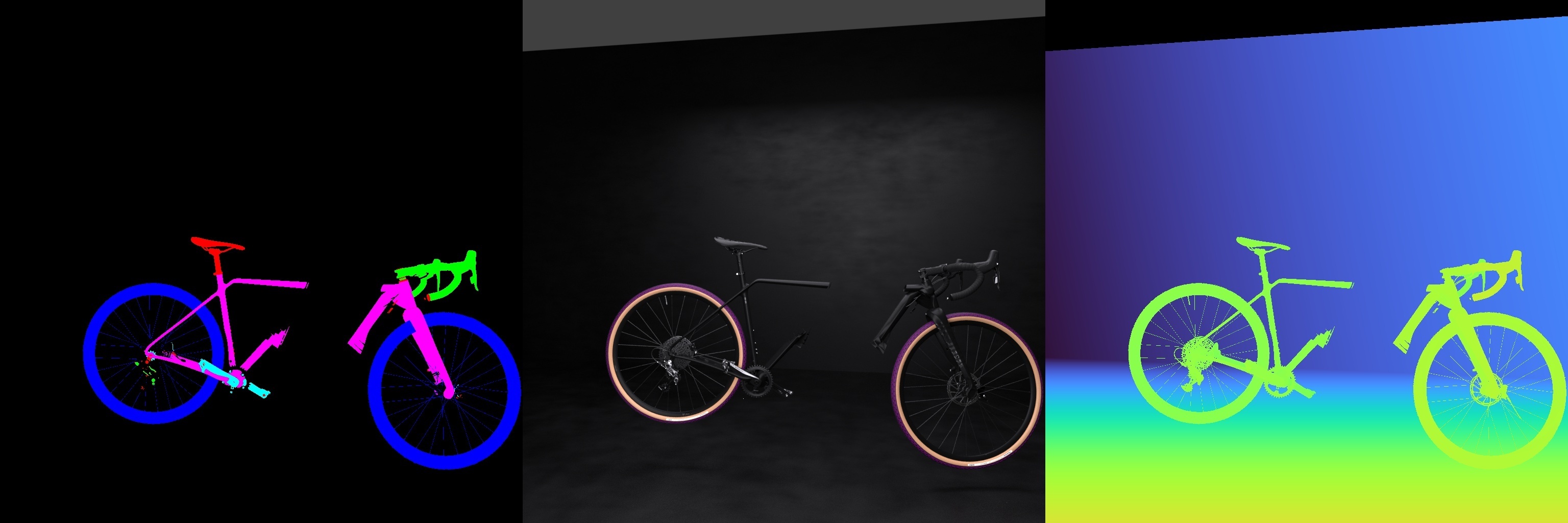}%
\label{fig:ex_4}}

\caption{Examples of the CGI pipeline auxiliary outputs. From right to left: segmentation, rendered image, and depth map.}
\label{fig:ex_add}
\end{figure}

For each ID, multiple images are generated by applying the following transformations:
\begin{enumerate}
    \item Mobile parts composition. This is accomplished by randomly performing one or more actions among: translating the seat and handlebar;  rotating the seat, handlebar, pedals, and wheels. The allowed range of movement for each model is set during the rig adaptation.
    \item  Dirt. A custom shader is used to randomly apply dirt in the form of mud or rust, with a predetermined probability.
    \item (Optional) Removing parts. The seat, pedals, handlebar and wheels can be removed with a predetermined probability. 
    \item (Optional) Damaging parts (frame excluded).  Parts of the bikes can be damaged, by selecting a deformation either from the wheels' library or from the pose library. 
    \item (Optional) Damaging the frame. The frame can be damaged either by picking a deformation from the pose library (Bent Frame) and/or by breaking it using the rail guides-boolean system (Broken frame). Hence, four possible damage categories are possible: normal frame, bent frame, broken frame, or bent \& broken frame. 
    \item  Point of View selection. The virtual camera position is chosen randomly within a given boundary, by randomly switching the visible side of the bike, as well as randomly adjusting the camera focal length within a parametrized range.
    \item Environment and Lighting. A combination of background and lighting setup is picked.
    \item Segmentation. The bike is segmented in the following classes: ``Front Wheel'', ``Rear Wheel'', ``Seat'', ``Crankset'', and ``Frame''. Segmentation was implemented using the \textit{bpycv} \footnote{\href{https://github.com/DIYer22/bpycv}{https://github.com/DIYer22/bpycv}} library. An example of segmentation is shown in \autoref{fig:ex_add}
 \end{enumerate}

\clearpage

\subsection{The \DATASET\ dataset}

In this section we provide additional details on the generated dataset to better illustrate the variety of models and damages/deformations included in the \DATASET{} dataset. 

\subsubsection{3D Bike models}

\begin{figure}[h]
\begin{tabular}{cccc}
\subfloat[becane]{\includegraphics[width=.22\textwidth]{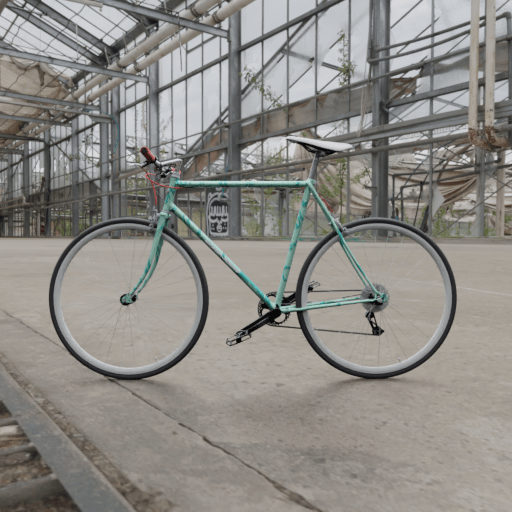}} &
\subfloat[btwin]{\includegraphics[width=.22\textwidth]{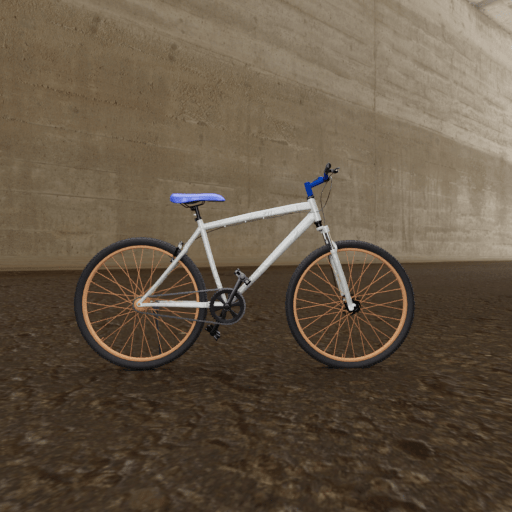}} &
\subfloat[croad]{\includegraphics[width=.22\textwidth]{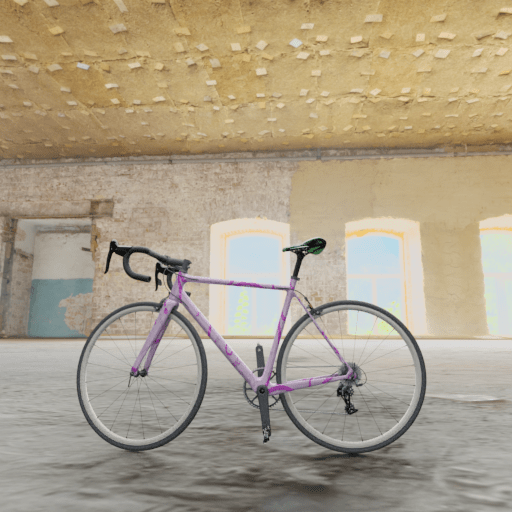}} &
\subfloat[domane]{\includegraphics[width=.22\textwidth]{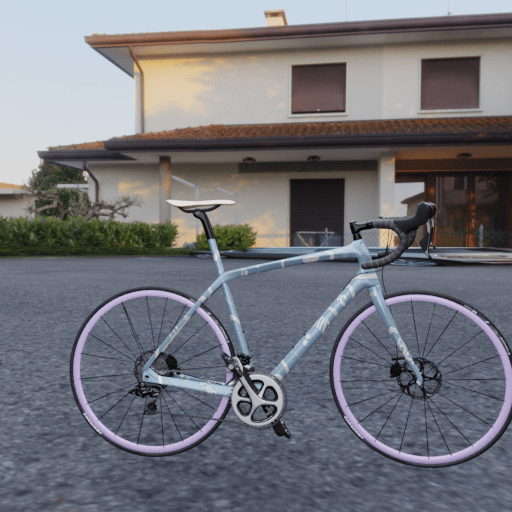}}\\
\subfloat[enduro]{\includegraphics[width=.22\textwidth]{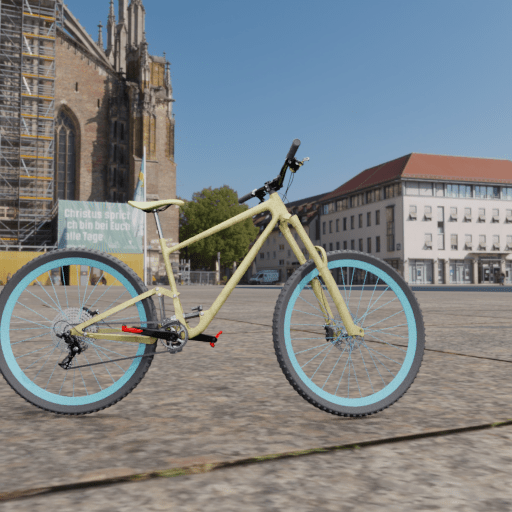}} &
\subfloat[freeride]{\includegraphics[width=.22\textwidth]{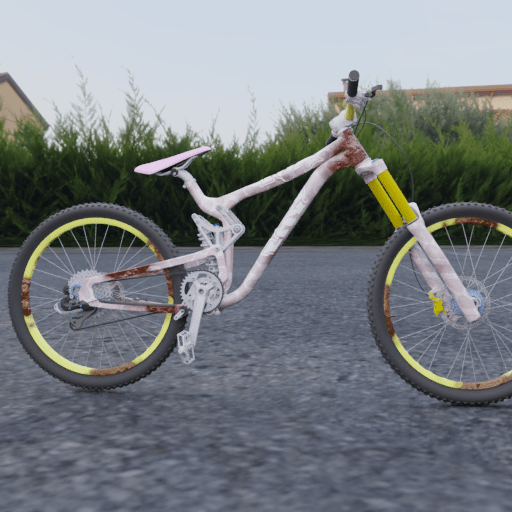}} &
\subfloat[g1]{\includegraphics[width=.22\textwidth]{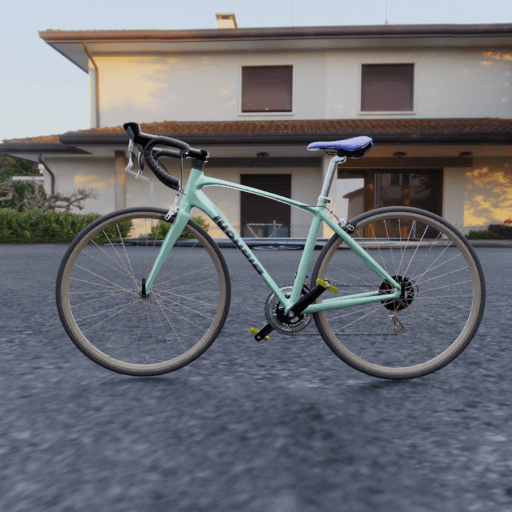}} &
\subfloat[gbike]{\includegraphics[width=.22\textwidth]{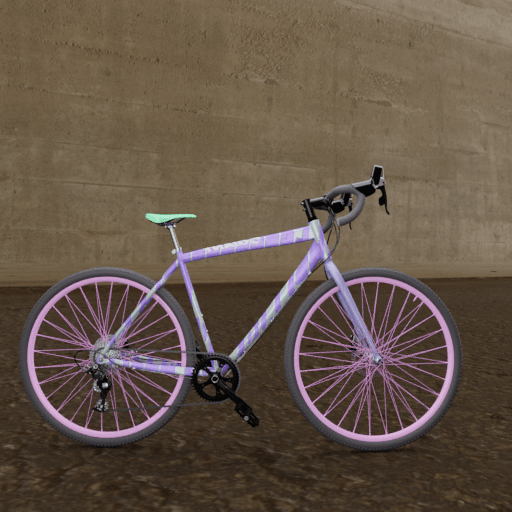}}\\
\subfloat[ghost]{\includegraphics[width=.22\textwidth]{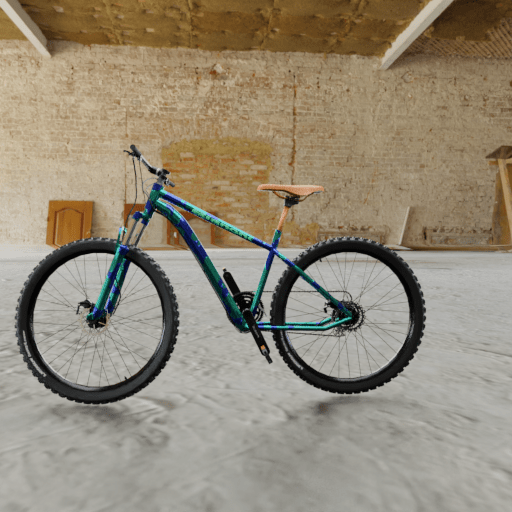}} &
\subfloat[holland]{\includegraphics[width=.22\textwidth]{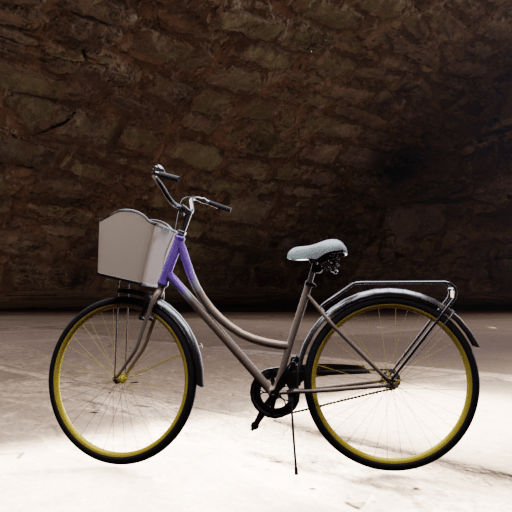}} &
\subfloat[huffy]{\includegraphics[width=.22\textwidth]{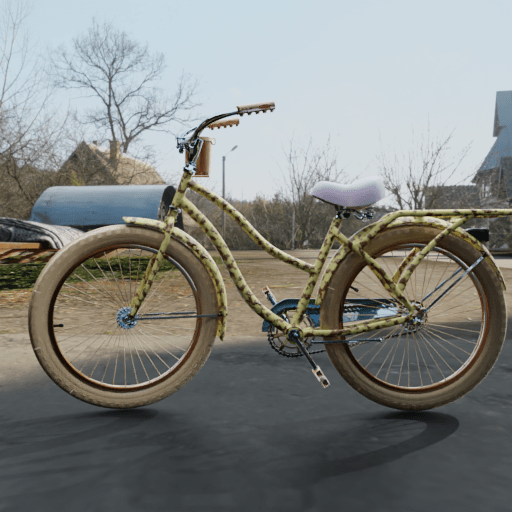}} &
\subfloat[kuota]{\includegraphics[width=.22\textwidth]{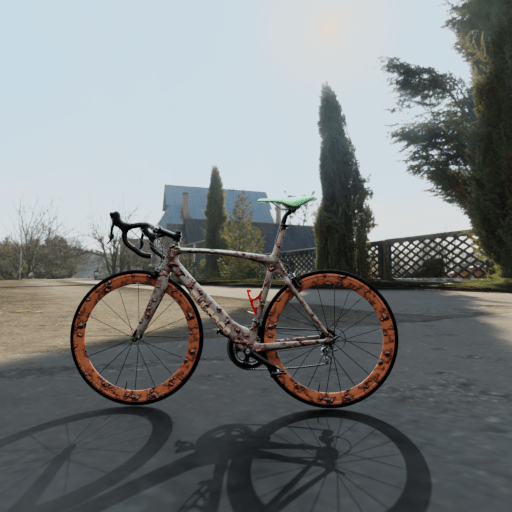}}\\
\subfloat[mfactory]{\includegraphics[width=.22\textwidth]{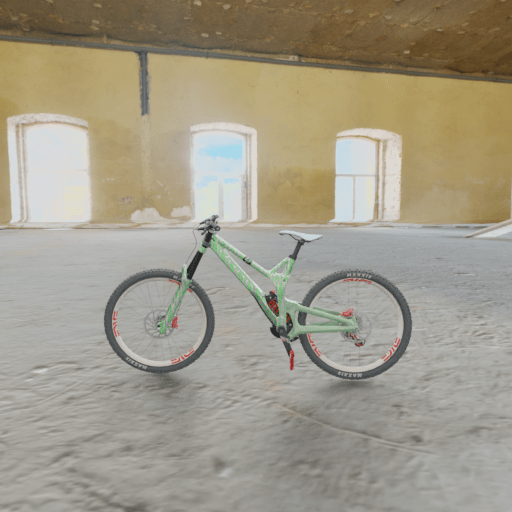}} &
\subfloat[mirage]{\includegraphics[width=.22\textwidth]{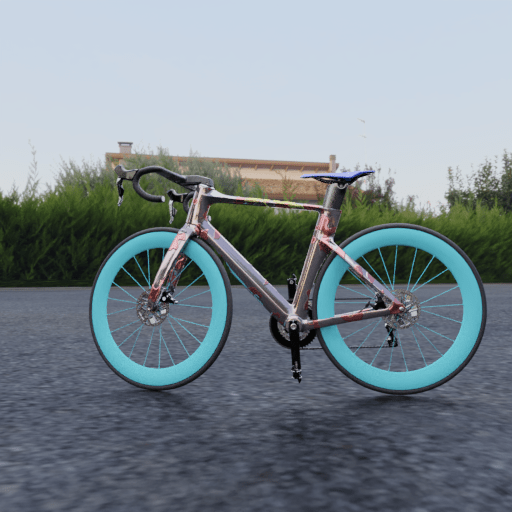}} &
\subfloat[oldbike]{\includegraphics[width=.22\textwidth]{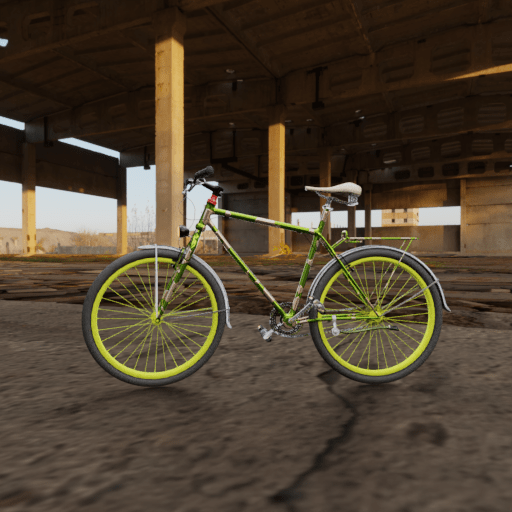}} &
\subfloat[rondo]{\includegraphics[width=.22\textwidth]{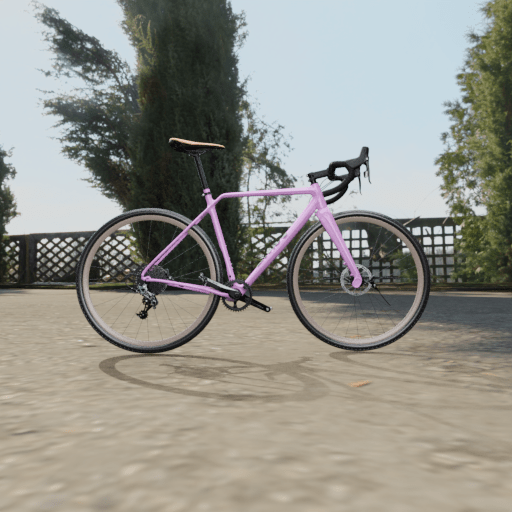}} \\
\subfloat[scalpel]{\includegraphics[width=.22\textwidth]{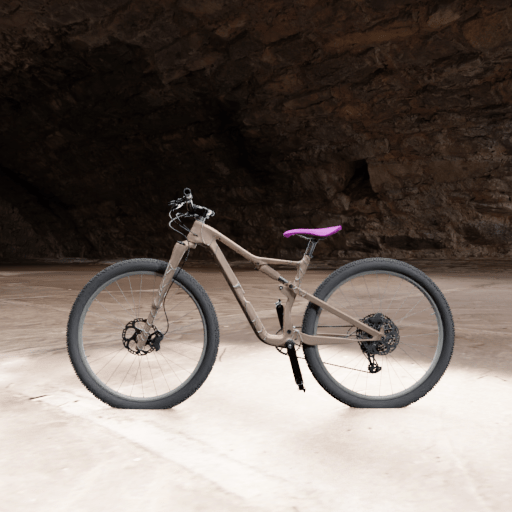}} &
\subfloat[verdona]{\includegraphics[width=.22\textwidth]{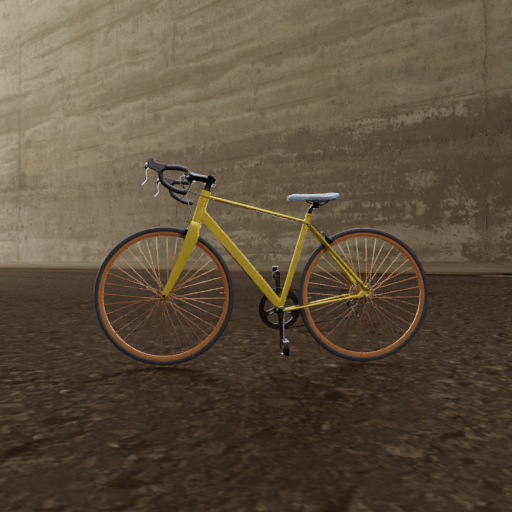}} &
\subfloat[vintage]{\includegraphics[width=.22\textwidth]{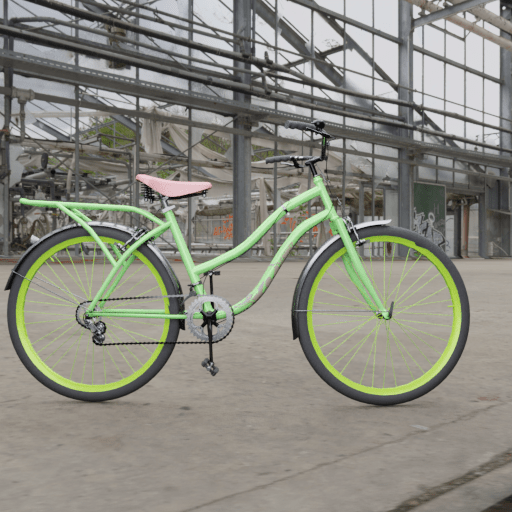}} &
\subfloat[wbike]{\includegraphics[width=.22\textwidth]{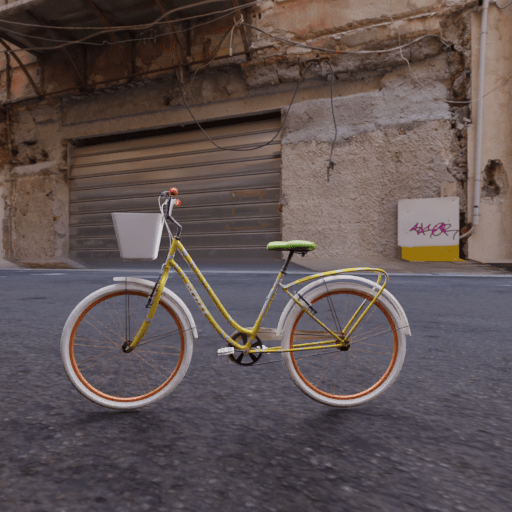}} 
\end{tabular}
\caption{Examples of each model used to generate the dataset.}
\label{fig:examples}
\end{figure}

\DATASET{} contains images generated from 20 3D bike models retrieved from dedicated marketplaces. It includes several variants of popular bikes such as Road, Cruiser and MTB. In particular, it contains 6 MTB, 1 Enduro, 6 Road, 1 Circuit, 1 Gravel, and 5 Cruiser. The list of bike models per category is illustrated in \autoref{tab:modelsplit}: each bike model was assigned to either the training, validation, or (stress) test set. Examples of renderings from each model are shown in \autoref{fig:examples}.


\begin{table}[htb]
\centering
\caption{Bike model distribution across \DATASET\ in the training, validation and test set. Models marked with (*) are shared between Train and Validation.}
\label{tab:modelsplit}
\resizebox{0.7\textwidth}{!}{%
\begin{tabular}{|c|ccc|}
\cline{1-4}
\multicolumn{1}{|c|}{\textbf{Category}}  & \multicolumn{1}{c|}{\textbf{Train}} &
\multicolumn{1}{c|}{\textbf{Validation}} &
\multicolumn{1}{c|}{\textbf{Test}} \\ 
 \hline
\multicolumn{1}{|c|}{MTB}         & mfactory & \multicolumn{1}{|c|}{becane}    &  - \\
\multicolumn{1}{|c|}{}        		  & ghost  & \multicolumn{1}{|c|}{btwin} &  \\ 
\multicolumn{1}{|c|}{}         & \multicolumn{2}{c|}{freeride*}   &   \\
\multicolumn{1}{|c|}{}         & \multicolumn{2}{c|}{scalpel*}    &   \\  \hline
\multicolumn{1}{|c|}{Road}        & rondo & \multicolumn{1}{|c|}{croad} & -  \\ 
\multicolumn{1}{|c|}{}         & verdona & \multicolumn{1}{|c|}{}  &   \\ 
\multicolumn{1}{|c|}{}         & ghost & \multicolumn{1}{|c|}{}  &   \\ 
\multicolumn{1}{|c|}{}        & \multicolumn{2}{c|}{domane*} &   \\
\multicolumn{1}{|c|}{}        & \multicolumn{2}{c|}{g1*} &   \\
\multicolumn{1}{|c|}{}        & \multicolumn{2}{c|}{kuota*} &   \\ \hline
\multicolumn{1}{|c|}{Cruiser}        & oldbike & \multicolumn{1}{|c|}{} & -     \\
\multicolumn{1}{|c|}{}        & \multicolumn{2}{c|}{holland*} &   \\
\multicolumn{1}{|c|}{}        & \multicolumn{2}{c|}{huffy*} &   \\
\multicolumn{1}{|c|}{}        & \multicolumn{2}{c|}{vintage*} &   \\
\multicolumn{1}{|c|}{}        & \multicolumn{2}{c|}{wbike*} &   \\\hline
\multicolumn{1}{|c|}{Enduro}      & \multicolumn{2}{c|}{-} 	&  enduro  \\ \hline
\multicolumn{1}{|c|}{Circuit}        &\multicolumn{2}{c|}{-}  &   mirage  \\ \hline
\multicolumn{1}{|c|}{Gravel} 		& \multicolumn{2}{c|}{-}  &      gbike     \\ \hline
\end{tabular}%
}

\end{table}

\clearpage

\subsubsection{Damage distribution and examples}

\begin{figure}[tb]
    \centering
    \subfloat{\includegraphics[width=0.48\textwidth]{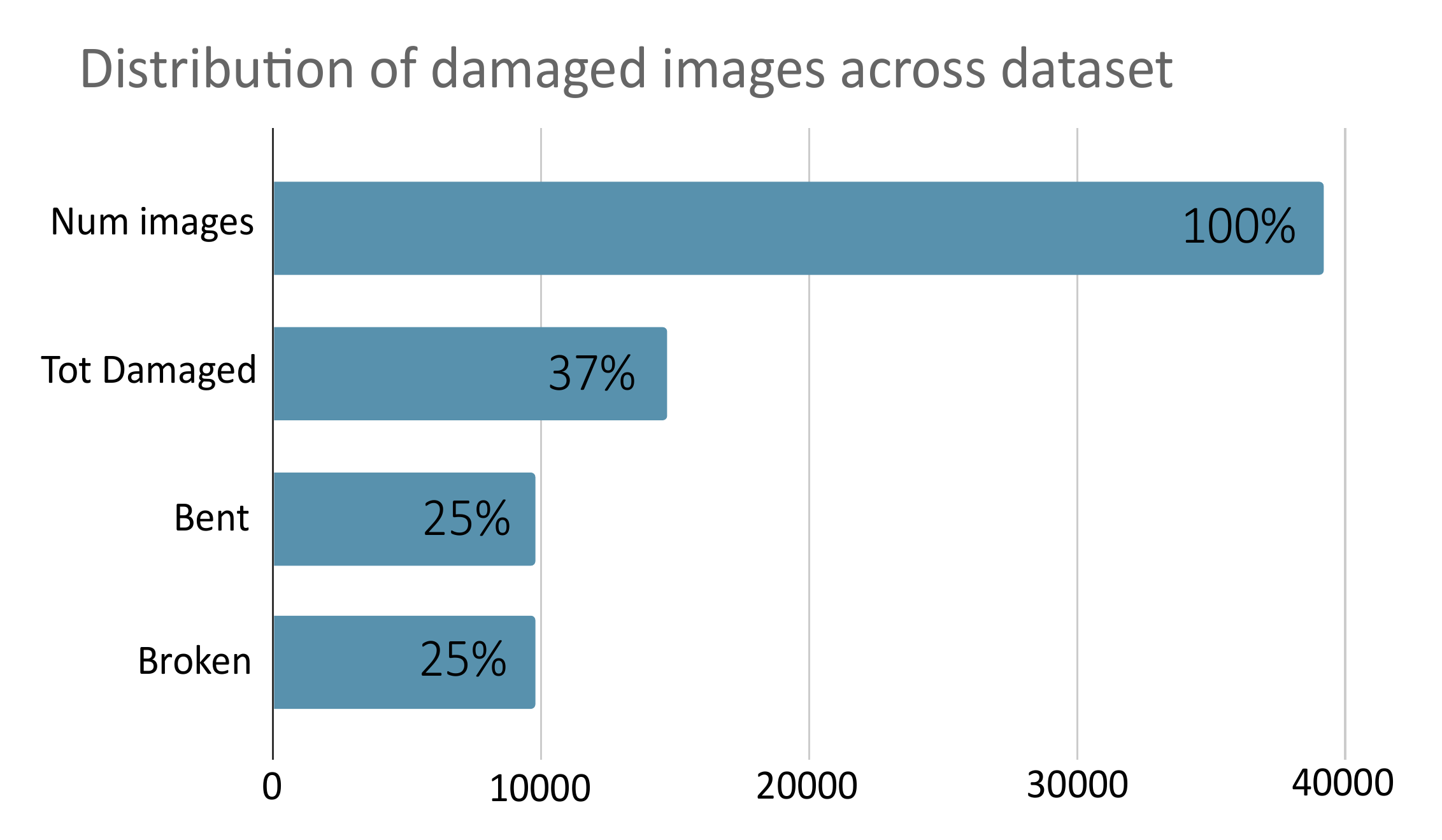}
    \label{fig:f1}}
    \hfill
    \subfloat{\includegraphics[width=0.48\textwidth]{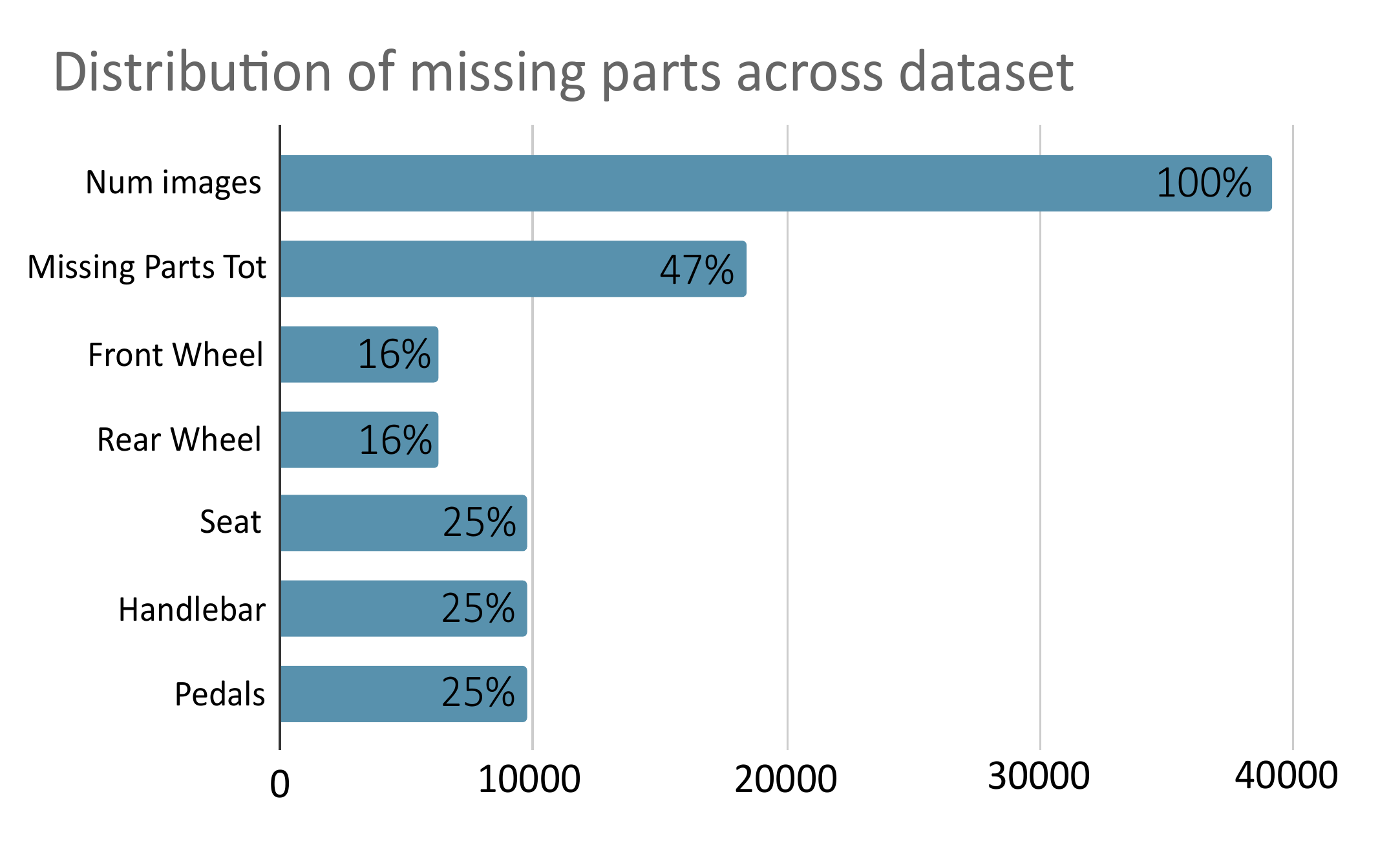}
    \label{fig:f2}}
    \caption{Distribution of damages and missing parts across the synthethic dataset.}
    \label{fig:distributions}
\end{figure}

\begin{figure}[!tbh]
                \centering
                \subfloat{\includegraphics[width=0.4\textwidth]{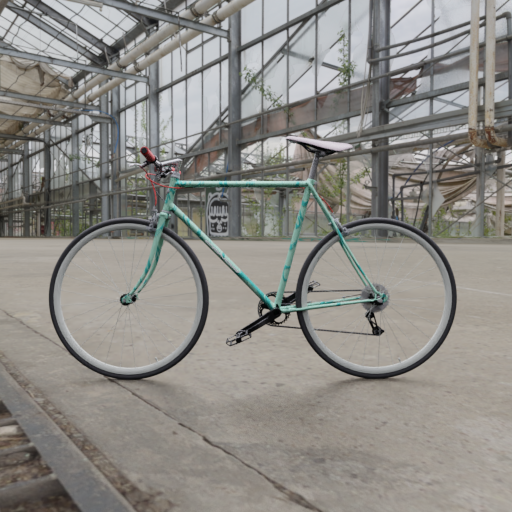}\label{fig:f1}}
                \hfill
                \subfloat{\includegraphics[width=0.4\textwidth]{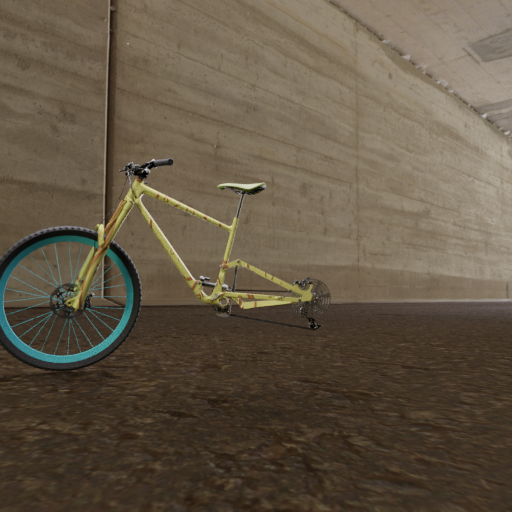}\label{fig:f2}}
                \caption{Examples of bike renderings without damages (to the frame). }
                \label{figdmg0}
\end{figure}

\begin{figure}[!tbh]
                \centering
                \subfloat{\includegraphics[width=0.4\textwidth]{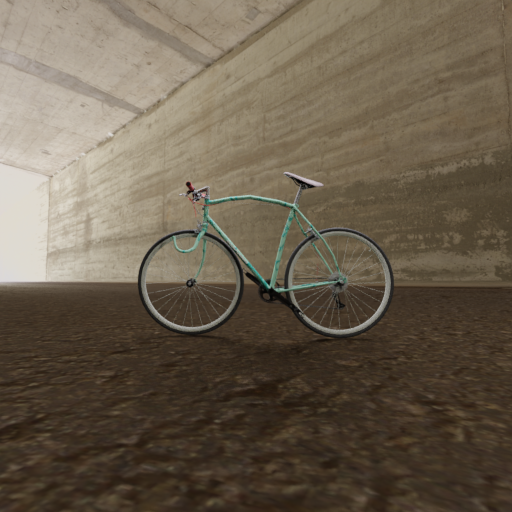}\label{fig:f1}}
                \hfill
                \subfloat{\includegraphics[width=0.4\textwidth]{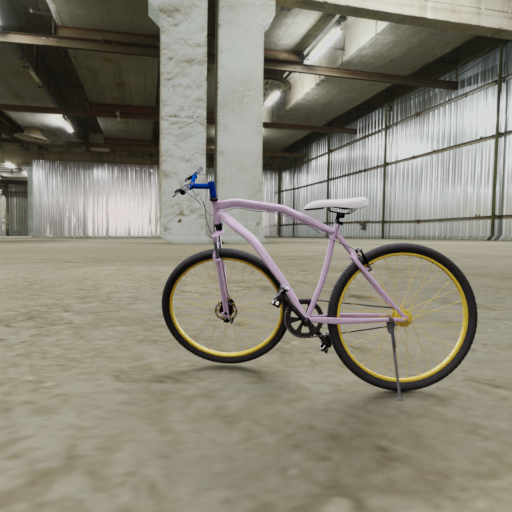}\label{fig:f2}}
                \caption{Examples of bike renderings with bent frames. }
                \label{figdmg1}
\end{figure}

\begin{figure}[!tbh]
                \centering
                \subfloat{\includegraphics[width=0.4\textwidth]{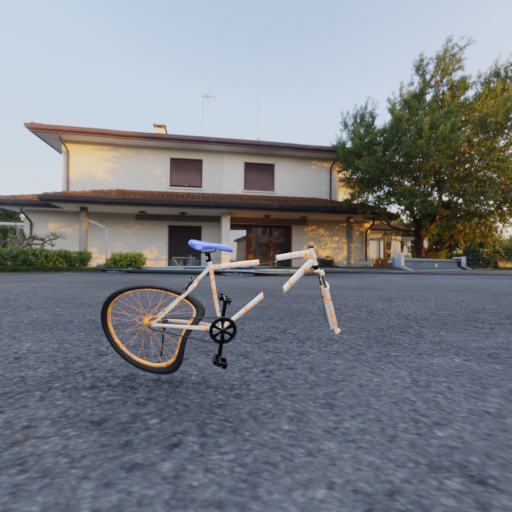}\label{fig:f1}}
                \hfill
                \subfloat{\includegraphics[width=0.4\textwidth]{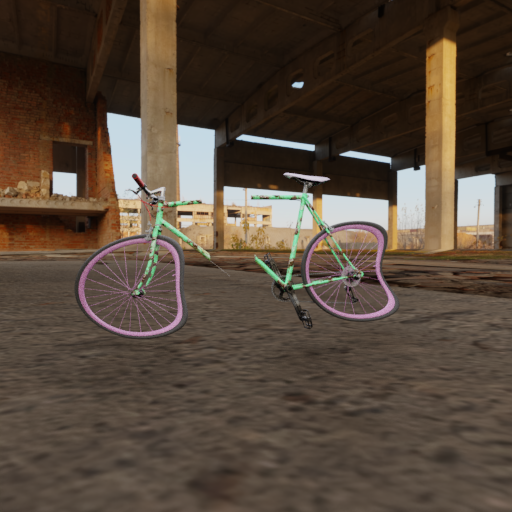}\label{fig:f2}}
                \caption{Examples of bike renderings with broken frames. }
                \label{figdmg2}
\end{figure}

\begin{figure}[!tbh]
                \centering
                \subfloat{\includegraphics[width=0.4\textwidth]{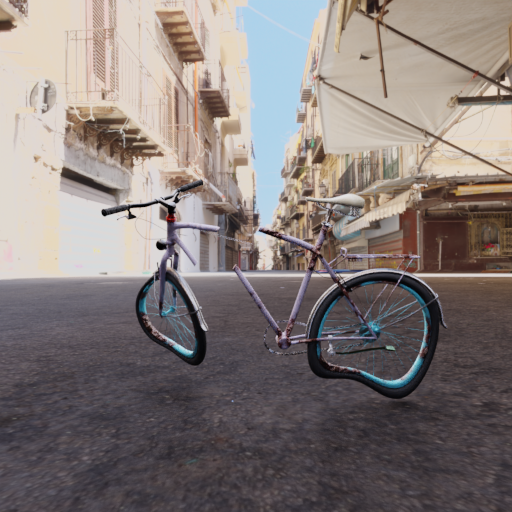}\label{fig:f1}}
                \hfill
                \subfloat{\includegraphics[width=0.4\textwidth]{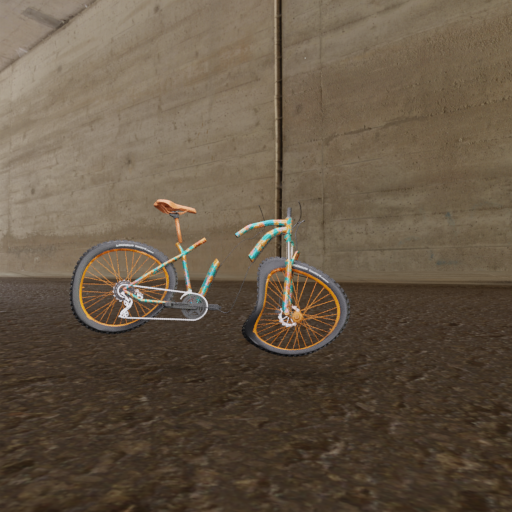}\label{fig:f2}}
                \caption{Examples of bike renderings with bent and broken frames.}
                \label{figdmg3}
\end{figure}

As illustrated in Section 3, in \DATASET{} 50\% of the images are generated ``before'' and 50\%  ``after'' a damage occurs.  ``Before'' bikes have a 25\% probability of being dirty, while  ``after'' bikes have a 50\% chance.  ``After'' bikes are further divided into 25\% undamaged and 75\% bent, broken or both. As a result,  37\% of the total images are damaged (see \autoref{fig:distributions}). Examples of damaged and undamaged synthetic bike renderings are shown in \autoref{figdmg0}, \autoref{figdmg1}, \autoref{figdmg2} and \autoref{figdmg3}.


\clearpage

Moreover, we set additional labels for each image according to the missing parts of the bike, namely: Front Wheel, Rear Wheel, Seat, Handlebar, Pedals. In the annotations, missing parts are represented by a One-hot vector encoding, where each vector value indicates if the corresponding part is present (0) or
not (1), as exemplified in Fig. \ref{fig:onehot_example}.

\begin{figure}[h]
    \centering
    \includegraphics[width=0.6\textwidth]{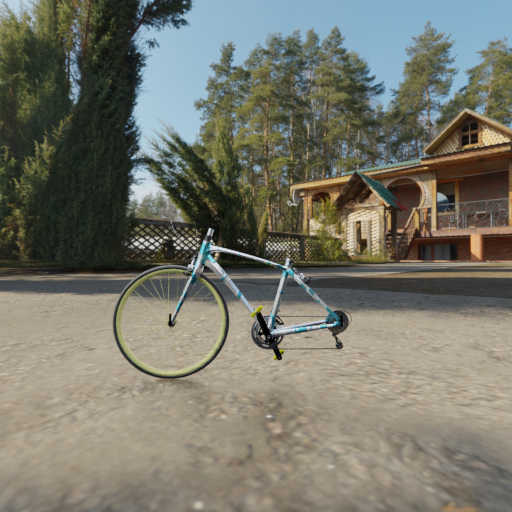}
    \caption{One-hot encoding example: the illustrated image only misses the ``Rear Wheel'', ``Seat'' and ``Handlebar'' parts, hence it has been labeled as ``01110''.}
    \label{fig:onehot_example}
\end{figure}

\subsection{Real dataset acquisition: additional details.}
In this section, we provide additional details on how the real dataset was assembled and annotated.

\subsubsection{Delft Bikes.}
The DelftBikes dataset\footnote{\href{https://data.4tu.nl/articles/dataset/DelftBikes_data_underlying_the_publication_Hallucination_In_Object_Detection-A_Study_In_Visual_Part_Verification/14866116}{https://data.4tu.nl/articles/dataset/DelftBikes\_data\_underlying\_the\_publication\_Hallucination\_In\_Object\_Detection\-A\_Study\_In\_Visual\_Part\_Verification/14866116}.} was originally designed to study whether deep neural networks could hallucinate missing parts in objects. It contains 10,000 bike images with 22 densely annotated parts for each bike. All part locations and part states (i.e., missing, intact, damaged, occluded) are explicitly annotated.

Specifically, we retained only the images from the DelftBikes training set with complete annotations (for some images missing parts annotations were not available), for a total of 8,000 images. Then, we translated the Delftbikes annotations to be compatible with the synthetic dataset annotations, as follows:
\begin{itemize}
    \item For Front Wheel, Rear Wheel and Seat, we labeled the part as missing if the corresponding part was labeled in the same way (i.e., object state class = missing) in the Delftbikes dataset.
    \item For Handlebar, we labeled the part as missing if all parts belonging to the group \{back handle, front handle, back hand break, front hand break, steer\} were also labeled in the same way (i.e., object state class = missing) in the Delftbikes dataset.
    \item For Pedals, we labeled the part as missing if both parts in the group \{front\_pedal, back\_pedal\} were also labeled in the same way (i.e, object state class = 2) in the Delftbikes dataset.
\end{itemize}

None of the bike instances in the DelftBikes dataset presented damages to the frame. 
\subsubsection{Web scraping details}

We collected samples of real damaged bikes by querying popular search engines (i.e., Google, Bing) and online forums (i.e., Reddit and other dedicated forums). We used different keywords (i.e., ``damaged bike'', ``bici danneggiata'', etc.) in different languages (i.e., English, Italian, Spanish, French, etc.) in order to increase the number of  matches. In particular, we selected countries with higher bike usage like the Netherlands and Denmark. Synonyms of damage were searched to amplify the number of returned images (for instance, ``broken bike'' and ``damaged bike'' produce different search results). Additional images of normal bikes were retrieved from second-hand e-commerce sites. 

For each scraped image, the origin URL has been serialized as a source reference. The results have then been pruned from unrelated (e.g., excluding images about bike helmets, cycling suits, etc.) and duplicated images by hand and by means of automatic de-duplication techniques, respectively. In particular, we chose a de-duplication technique based on pre-trained CNNs, which marks as duplicated images those with a pairwise similarity score above a given threshold value (experimentally set to 85\%). 

\subsubsection{Labelling.}

All images were manually labeled indicating the damage type and missing parts. Labels were assigned as uniformly as possible to the synthetic dataset. Concerning damage labeling, we set four different labels based on the type of damage present on the bike frame:
\begin{itemize}
    \item normal: the bike frame is intact, regardless of the condition of the other parts of the bike (e.g., missing parts, damaged wheels, damaged seat).

    \item bent: the bike frame is bent or presents damage, but it is broken in multiple pieces.

    \item broken: the bike frame is broken and clearly divided in pieces, and each piece does not present any bending.

    \item bent \& broken: the bike frame is broken and the frame pieces show signs of bending.

\end{itemize}

\begin{figure}[tb]
    \centering
     \begin{subfigure}[b]{0.48\textwidth}
         \centering
         \includegraphics[width=\textwidth]{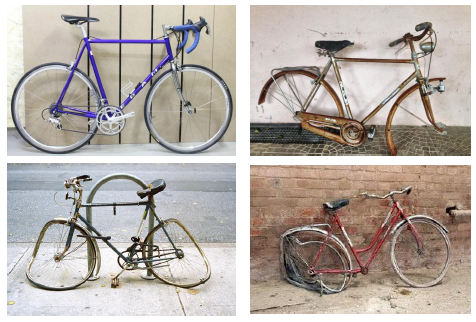}
         \caption{Normal}
         
     \end{subfigure}
     \hfill
     \begin{subfigure}[b]{0.48\textwidth}
         \centering
         \includegraphics[width=\textwidth]{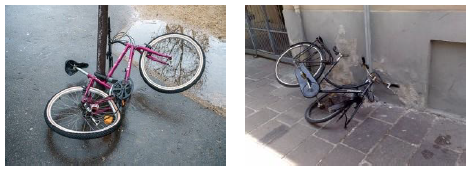}
         \caption{Bent}
       
     \end{subfigure}
     \\  
     \begin{subfigure}[b]{0.48\textwidth}
         \centering
         \includegraphics[width=\textwidth]{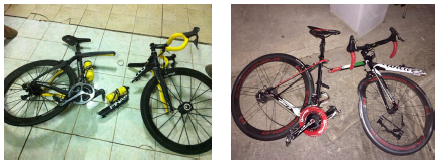}
         \caption{Broken}
   
     \end{subfigure}
     \hfill
        \begin{subfigure}[b]{0.48\textwidth}
         \centering
         \includegraphics[width=\textwidth]{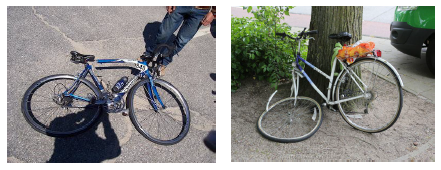}
         \caption{Bent \& broken}

     \end{subfigure}

        \caption{Examples of real images with normal (a), bent (b), broken (c) and bent \& broken (d) frames. }
        \label{fig:realexamples}
\end{figure}

For missing parts, we follow the same convention of the synthetic dataset and set additional labels for the following parts: Front Wheel, Rear Wheel, Seat, Handlebar, Pedals. Missing parts are represented by a One-hot vector encoding, where each vector value indicates if the corresponding part is present (0) or
not (1), as exemplified in Fig. \ref{fig:onehot_example}.

\begin{figure}[h]
    \centering
    \includegraphics[width=0.5\textwidth]{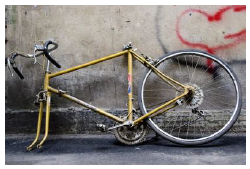}
    \caption{Example of real image with missing parts: the illustrated image only misses the ``Front wheel'' and ``Seat'' parts, hence it has been labeled with ``10100''.}
    \label{fig:onehot_example}
\end{figure}

\section{Experimental settings}

\textit{classification \CLS{} embedding}, which encodes image global features, is prepended to the sequence of $N$ \textit{patch tokens}. Each token is encoded by the combination (sum) of the corresponding patch embedding, learnable positional embedding, and SIE embeddings. The $N+1$ input tokens, inclusive of the \CLS{} token for a total size of $[1, N+1, 768]$, are then input into the transformer backbone. 
    
\begin{itemize}     
    \item \textbf{Shared ViT Network}: a ViT-like structure including $L-1$ layers ($L = 12$) is used as a shared backbone, whose output is then passed to each task-dedicated branch, each including an additional separate $L^{th}$ transformer layer; each layer attention module has 12 attention heads.
    
    \item \textbf{ReID global branch}: the ReID task is performed based on the  $[cls]$ token alone (which is a global representation of the image features). The token is first passed through a Batch Normalization (BN) layer, whose output is first used for the triplet loss calculation and then passed to a FC layer for performing the ID cross-entropy loss computation. 
    
    \item \textbf{Jigsaw Branch and Jigsaw Patch Module}: in the Jigsaw branch, the Jigsaw Patch Module (JPM) module is applied on the output of the $L-1$ shared transformer layers: first the $[cls]$ token is separated from the output of the $L-1$ layers, while the remaining part of the output, consisting only of the patch tokens, is randomly rearranged into four equally $N/4$ sized groups. Then, the previously extracted $[cls]$ token is added to each group so obtained, and each group is finally passed to $L^{th}$ transformer layer of the JPM branch; the output of the JPM branch is a set of classification tokens, one for each group. In the same way as in the global branch, each output \CLS{} token is passed through a corresponding BN layer, whose output is used for the triplet loss calculation and then passed to the corresponding FC layer for the ID cross-entropy loss. These loss components are added to the combined loss of the global branch to be minimized. In this way, the ReID model learns more discriminative parts and becomes more robust with respect to perturbations.
    
    \item \textbf{Damage branch}: like for the ReID task, the damage classification is performed on the \CLS{} token alone. The token is first passed through 7 different BN layers (one per head), and each output is passed to a corresponding FC layer, one for Bend frame classification, one for Broken frame classification, and one for each missing part classification. The scores and cross-entropy losses produced by each of these heads are then combined by weighted averaging for the final damage loss.  

\end{itemize}

\subsection{Domain adaptation}

\begin{figure}[ht]
    \centering
    \includegraphics[width=1\textwidth]{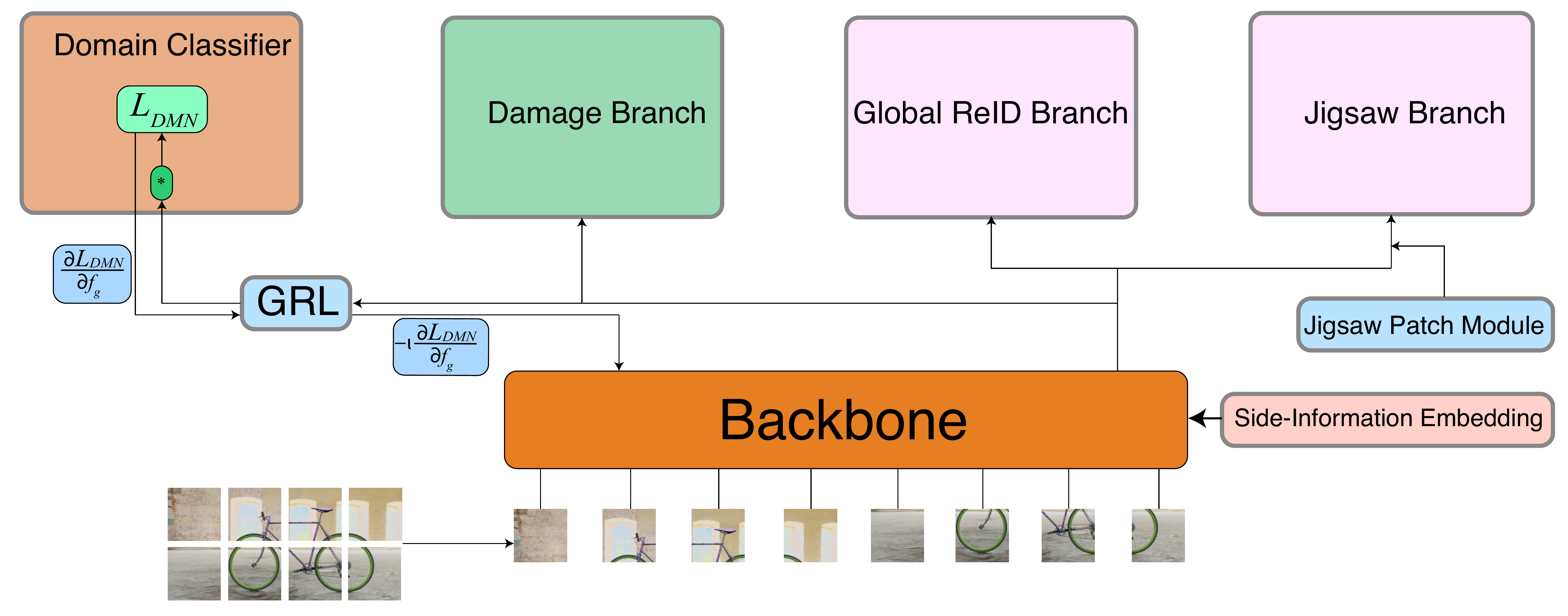}
    \caption{ \NETWORK\ architecture with the addition of DANN components.}
    \label{fig:TDReID_dann}
\end{figure}

\begin{figure}[ht]
    \centering
    \includegraphics[width=0.9\textwidth]{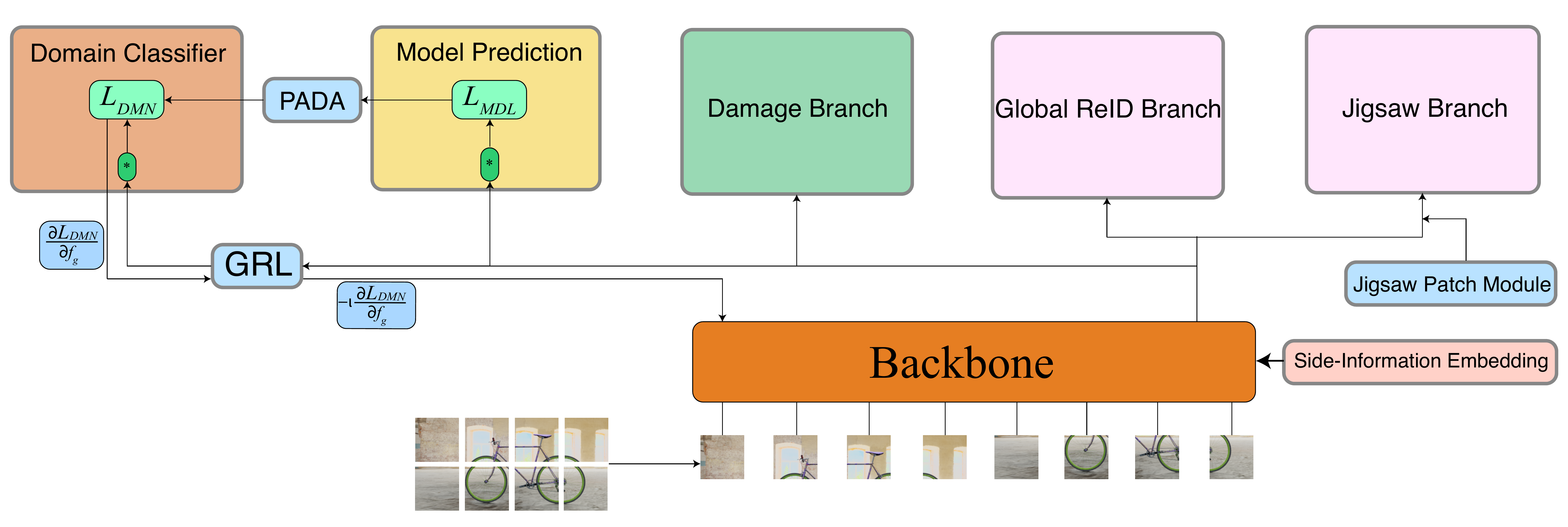}
    \caption{ \NETWORK\ architecture with the addition of the PADA module.}
    \label{fig:TDReID_pada}
\end{figure}

The resulting architecture configurations after the addition of DANN and PADA are depicted in Fig.\ref{fig:TDReID_dann} and Fig.\ref{fig:TDReID_pada}, respectively.

Parameters used for domain adaptation are:
\begin{itemize}
    \item $ \theta =1.0$ as weight for the domain discriminator loss $mathcal{L}_{dmn}$ (\autoref{eq:padadann_loss}).
    \item $ \delta =1.0$ as weight for the model classification loss $\mathcal{L}_{mdl}$, when PADA is active, otherwise 0 (\autoref{eq:padadann_loss}).
    \item Gradient Reversal Layer weight $ \iota =1.0$ in all DANN and PADA experiments except for the Base + Real + DANN experiment, in which $ \iota =10.0$ (\autoref{eq:padadann_loss}).
\end{itemize}

\begin{equation} \label{eq:padadann_loss}
   \mathcal{L}_{D\_tot} = \mathcal{L}_{D} + \theta \mathcal{L}_{dmn} + \delta \mathcal{L}_{mdl}  -\iota \frac{\partial \mathcal{L}_{DMN}}{\partial f_g}
\end{equation}

\section{Additional results}
\subsection{Retrieval examples}

\begin{figure}[h]
     \includegraphics[width=\textwidth]{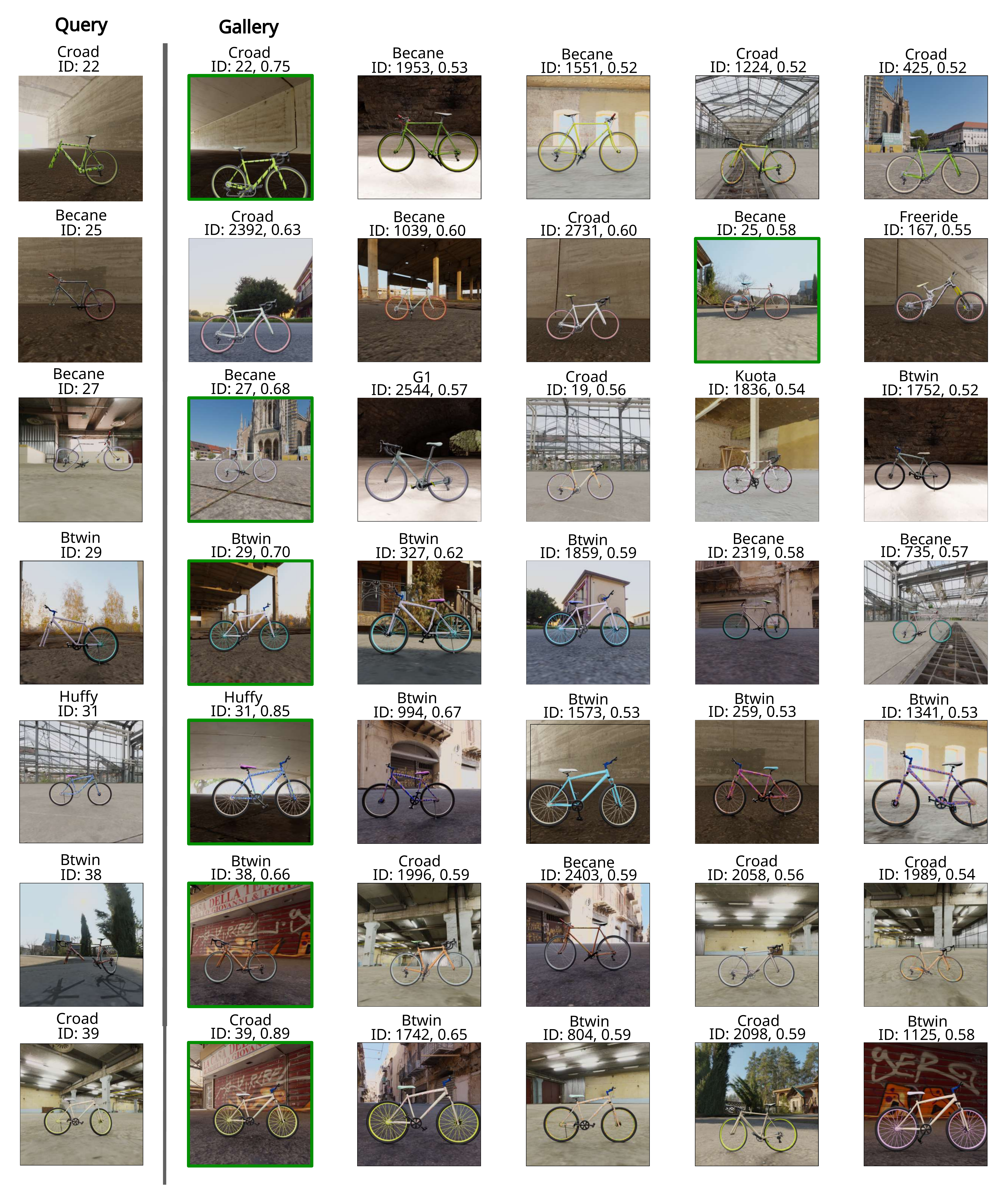}
    \caption{Retrieval results (Top-5 images) for the Baseline configuration, and corresponding model, bike ID, similarity scores. In most cases, the correct result (shown with a green border) is within the Top-5 predictions.}
    \label{fig:retrievals}
\end{figure}

\autoref{fig:retrievals} depicts some example predictions of \NETWORK{} on the ReID task. 

\clearpage

\subsection{Effect of the background on ReID and DD tasks}
\begin{figure}[tb!]
    \centering
    \begin{subfigure}[b]{\textwidth}
        \centering
        \includegraphics[width=\textwidth]{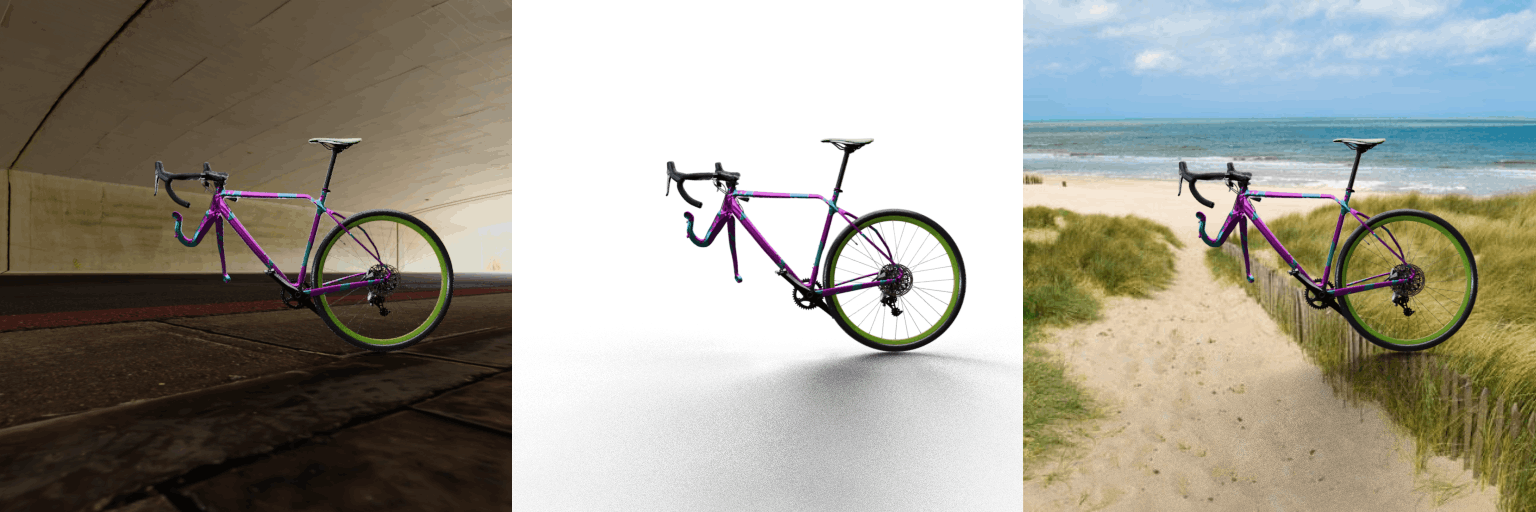}
    \end{subfigure}
    \begin{subfigure}[b]{\textwidth}
         \centering
         \includegraphics[width=\textwidth]{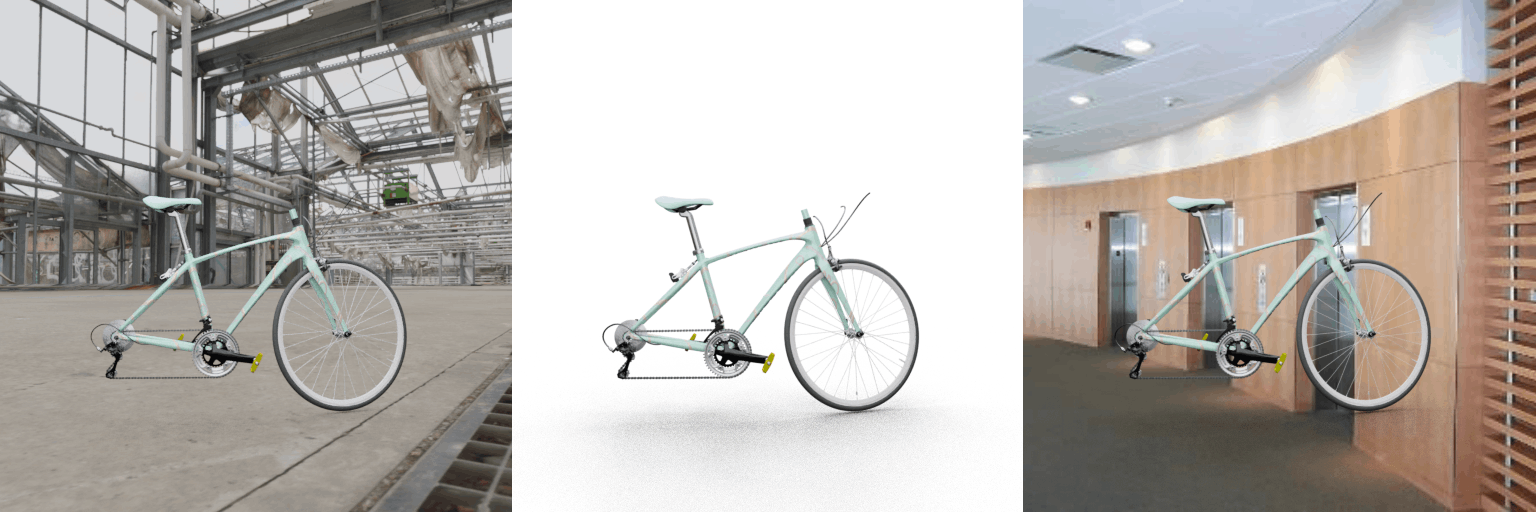}
    \end{subfigure}
    \begin{subfigure}[b]{\textwidth}
        \centering
        \includegraphics[width=\textwidth]{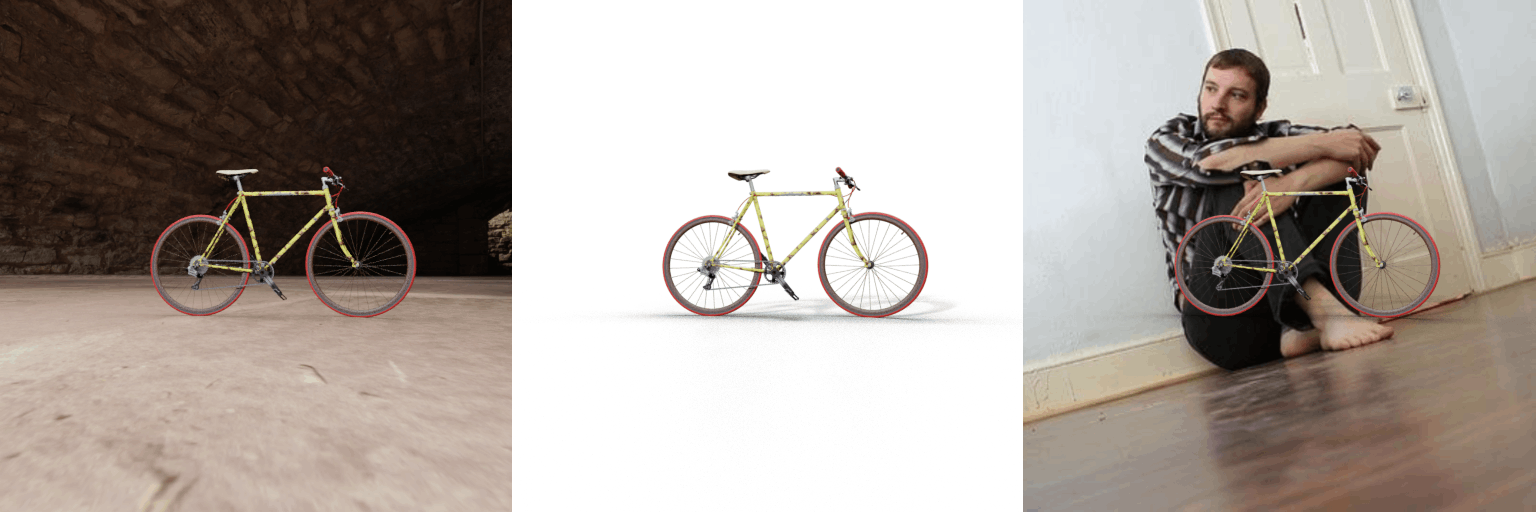}
    \end{subfigure}
    \caption{Examples of synthetic bike rendering placed against a 360 HDR background, a uniform background, and random scene from the Places365 dataset.}
    \label{fig:places365}
\end{figure}

In this section, we report additional experiments with different backgrounds to understand its effect on the ReID and DD tasks. Specifically, we compared three techniques: (i) the use of HDRI images, with random camera position, to generate the background, as detailed in Section 3; (ii) randomly picking an image from Places365 as background, and (iii) the use of a simple uniform background.  To generate samples with Places365 images as backgrounds, we leveraged the original pipeline to render an auxiliary image with a transparent background and overlay it over the photo taken from the dataset. Examples are shown in \autoref{fig:places365}. It should be noticed that the proposed pipeline leverages a limited number of 360° HDRI maps, and even if the proposed pipeline can generate a virtually infinite number of backgrounds by varying the bike position, camera and illumination, the backgrounds will be visually correlated. On the other hand, Places365 contains a much wider range of scenes, but since the bike is randomly positioned, the resulting blend is not always realistic, and the foreground and background are not as consistent as with HDRI maps. 

In \autoref{tab:res_back}, we evaluate the ability of \NETWORK{} to generalize to the real domain in the following transfer scenarios: $ \text{HDRI} \to \text{Real} $, $ \text{Places365} \to \text{Real} $ and $ \text{Uniform} \to \text{Real}$. We found moderately better results when employing the HDRI results, although when real images are available at training time, with HDRI yielding a marginal improvement over Places365. 

We further assess the ability to transfer across synthetic domains, specifically we evaluate the following scenarios: $ \text{Places 365} \to \text{HDRI}$ and $ \text{Uniform} \to \text{HDRI}$. We found that the network generalizes quite well across different strategies to insert the background, as long as it is not uniform. In the latter case, the performance significantly drops as the network is no longer able to separate the bike from the background. 

Based on these results, we conclude that the proposed pipeline contains sufficiently varied backgrounds, and the higher consistency improves the generalization capabilities.

\begin{table}[h]
\caption{\NETWORK{} performance on the validation set with different strategies to generate the background. The network was trained on synthetic data  except for $\dagger$ (labeled real images available at training time) and $\ddagger$ (unlabelled real images available at training time). }
\label{tab:res_back}
\resizebox{\textwidth}{!}{%

\begin{tabular}{c|cccccc|}
\cline{2-7}
& \multicolumn{6}{c|}{\cellcolor[HTML]{EFEFEF}\textbf{Validation}}  \\ 
\cline{2-7} 

& \multicolumn{2}{c|}{\textbf{Damage Detection}}             & \multicolumn{4}{c|}{\textbf{Re-identification (Synthetic)}} \\ 
\cline{2-7} 
 & \multicolumn{1}{c|}{\textbf{Real AUC}} &
  \multicolumn{1}{c|}{\textbf{Synthetic AUC}} & 
  \multicolumn{1}{c|}{\textbf{mAP}} &
  \multicolumn{1}{c|}{\textbf{CMC-1}} &
  \multicolumn{1}{c|}{\textbf{CMC-5}} &
  \textbf{CMC-10} \\ 
 \hline

\multicolumn{1}{|c|}{BG HDRI + Real\textsuperscript{$\dagger$}}        & \textbf{97.3 ± 2.2} & \multicolumn{1}{c|}{\textbf{91.4 ± 0.2}}          & \textbf{85.3 ± 0.2}   & \textbf{79.4 ± 0.1}           & \textbf{92.9 ± 0.4}          & \textbf{96.6 ± 0.4} \\ \hline
\multicolumn{1}{|c|}{BG Places365 + Real \textsuperscript{$\dagger$}}     & 96.3 ± 1.9 & \multicolumn{1}{c|}{90.4 ± 0.2}      &  85 ± 0    & 79.0 ± 0.4           & 92.8 ± 0.3          & 96.3 ± 0.2          \\
\multicolumn{1}{|c|}{BG Uniform + Real \textsuperscript{$\dagger$}}       & 95.2 ± 3.4 & \multicolumn{1}{c|}{87.4 ± 1.5}      & 48.5 ± 3.4    & 39.2 ± 1.9           & 59.4 ± 5.6          & 66.0 ± 5.7          \\ \hline
\end{tabular}%
}

\end{table}

\clearpage

\subsection{Additional explainability and t-SNE plot}
\begin{figure}[th!]
    \centering
     \begin{subfigure}[b]{0.2\textwidth}
         \centering
         \includegraphics[width=\textwidth]{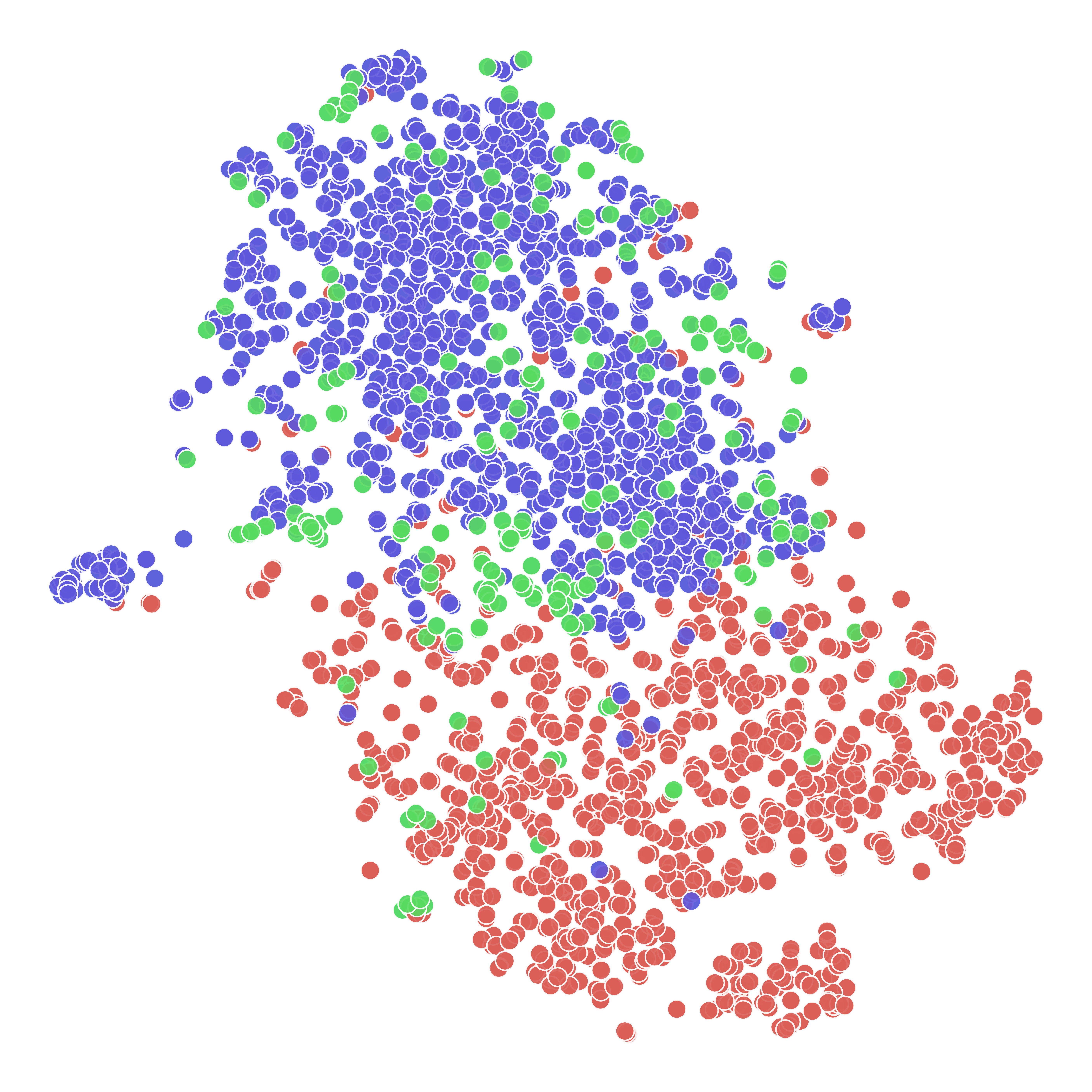}
         \caption{Baseline}
         \label{fig:bl_tsne}
     \end{subfigure}
     \hfill
     \begin{subfigure}[b]{0.2\textwidth}
         \centering
         \includegraphics[width=\textwidth]{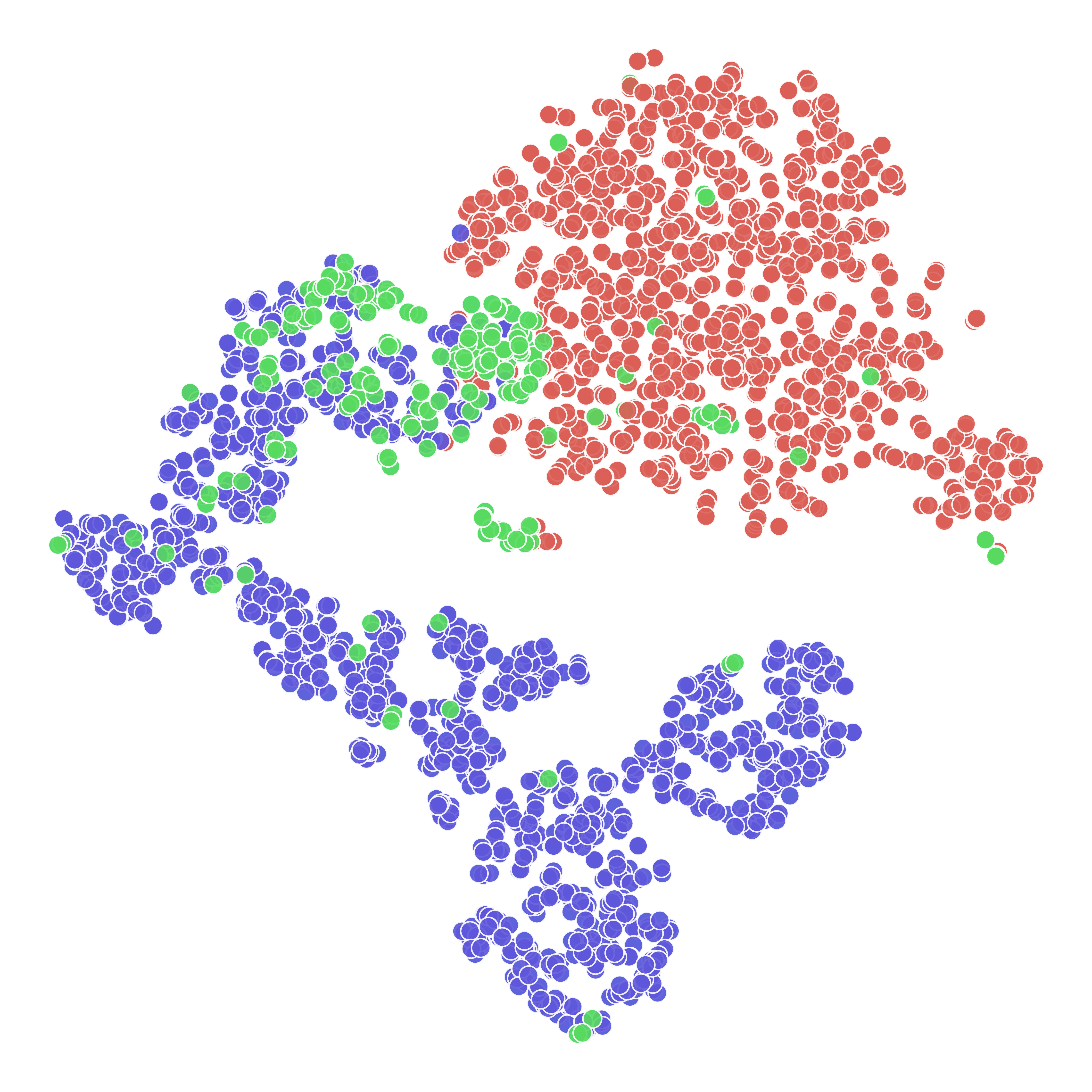}
         \caption{BL + Real}
         \label{fig:blreal_tsne}
     \end{subfigure}
     \hfill
     \begin{subfigure}[b]{0.2\textwidth}
         \centering
         \includegraphics[width=\textwidth]{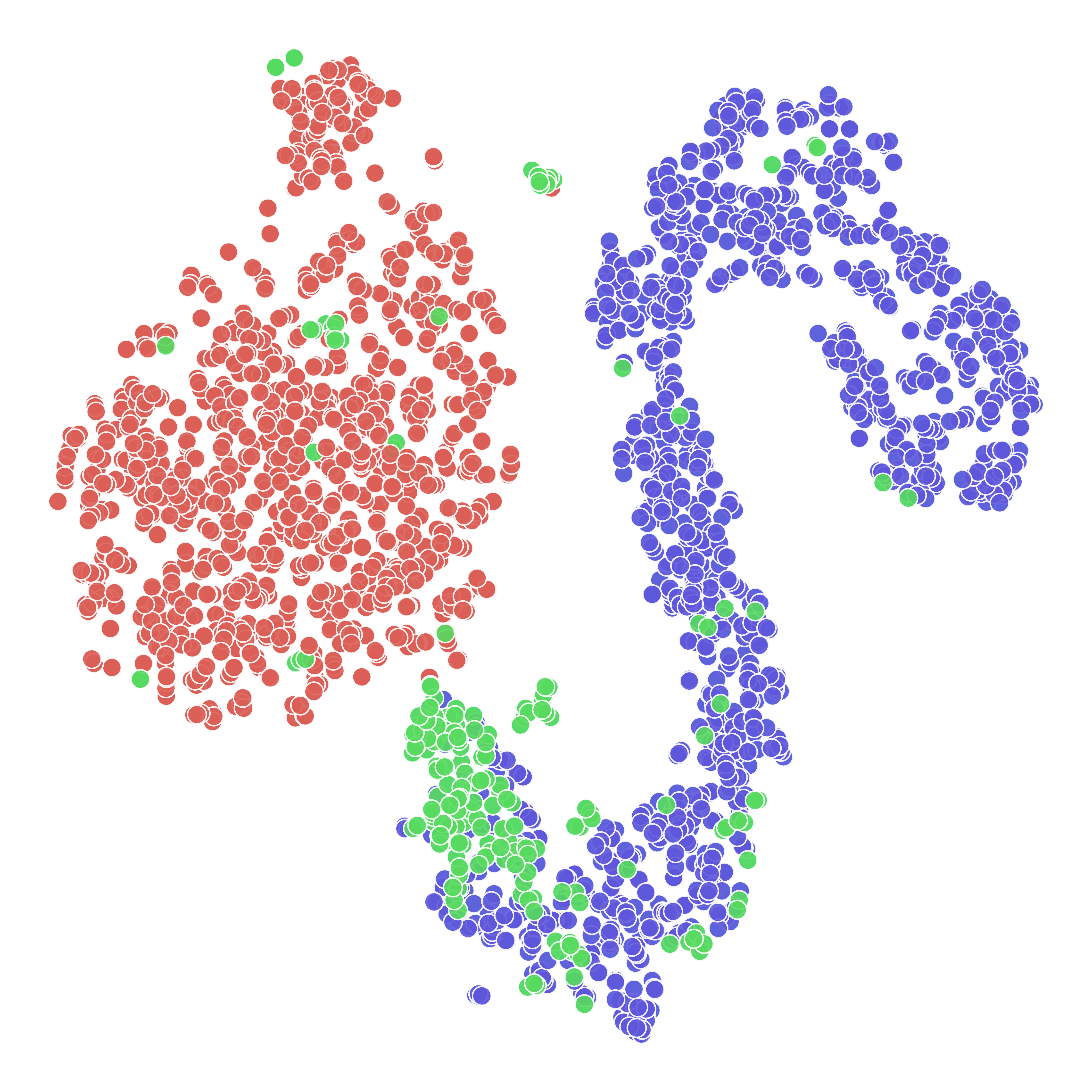}
         \caption{BL + R + DANN}
         \label{fig:dann_tsne}
     \end{subfigure}
     \hfill
     \begin{subfigure}[b]{0.2\textwidth}
         \centering
         \includegraphics[width=\textwidth]{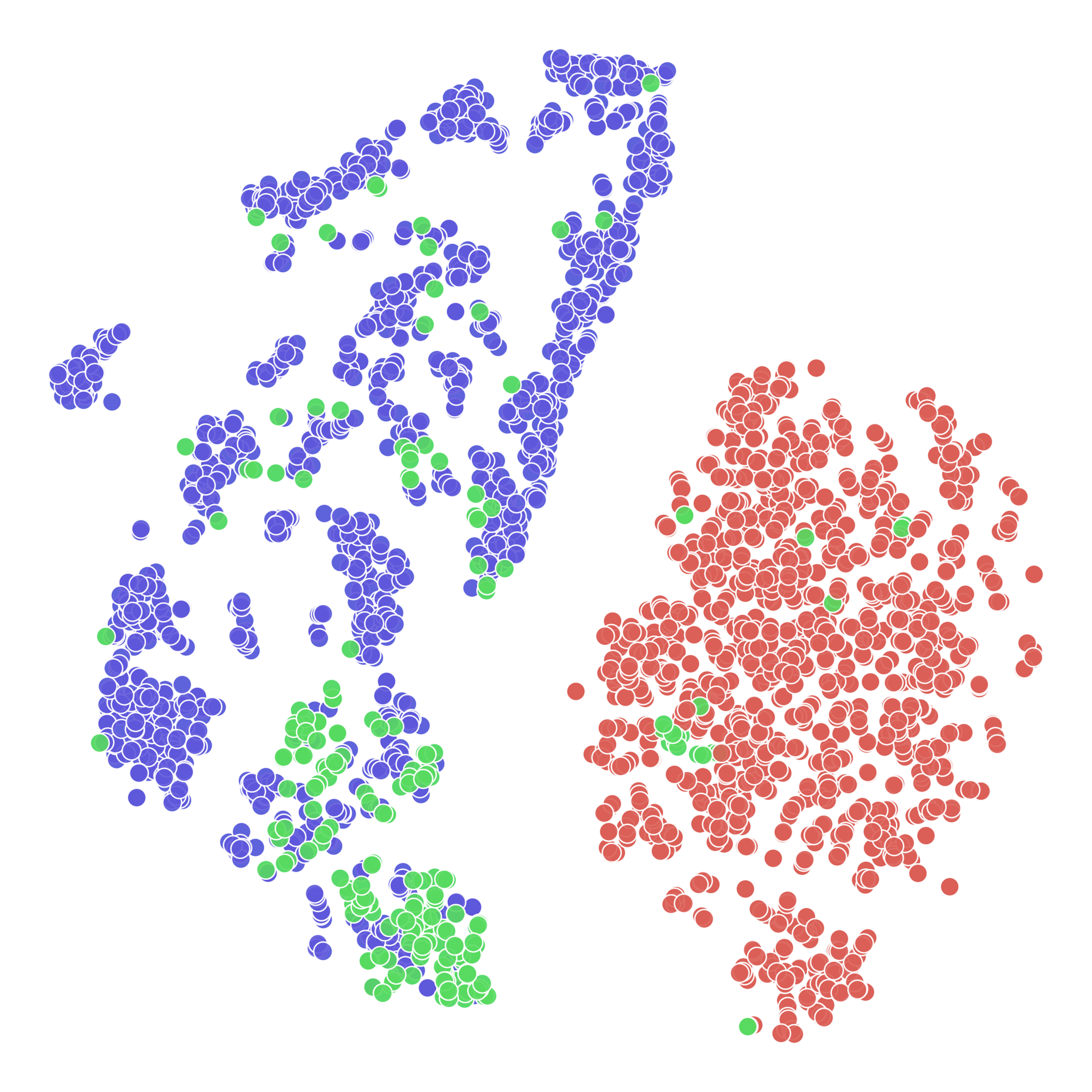}
         \caption{BL + R + PADA}
         \label{fig:PADA_tsne}
     \end{subfigure}
        \caption{t-SNE plot of the \CLS{} token extracted from the DD branch under different training regimes.  \textcolor{snsblue}{\textbullet{}} DelftBikes (real) \textcolor{snsgreen}{\textbullet} Web scraping (real) \textcolor{snsred}{\textbullet} \DATASET{} (synth) }
        \label{fig:tsne_dmg}
\end{figure}

The t-SNE plots of the \CLS{} token extracted from the backbone (\autoref{fig:tsne_dmg}) show partial overlap between the real and synthetic domains, and highlight how real images from various sources yield very different distributions. 

\begin{figure}[th!]
    \begin{center}
         \begin{subfigure}[b]{0.2\textwidth}
             \centering
    \includegraphics[width=\textwidth]{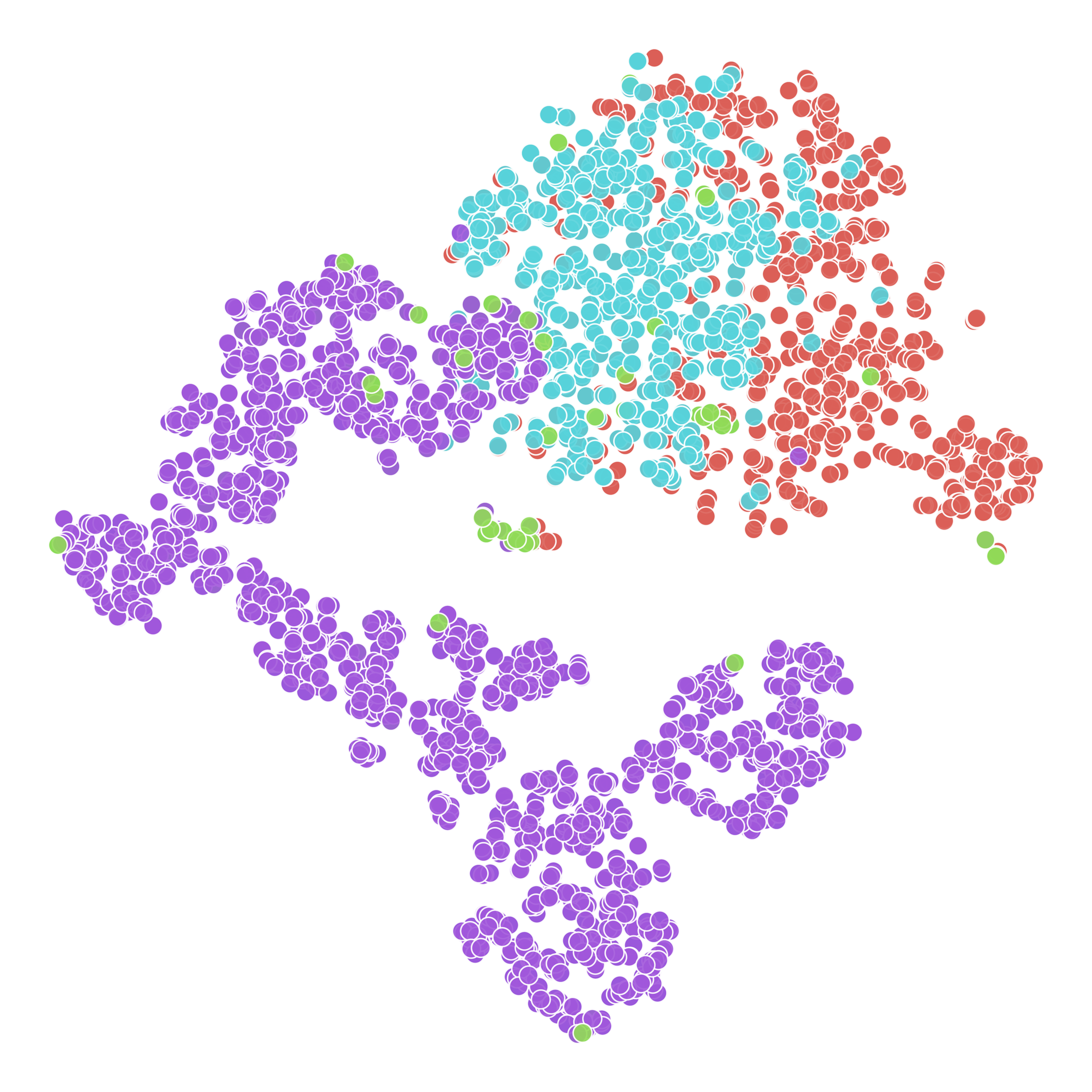}
             \caption{Damage branch}
             \label{fig:domain_shift_bb}
         \end{subfigure}
         \begin{subfigure}[b]{0.2\textwidth}
             \centering
             \includegraphics[width=\textwidth]{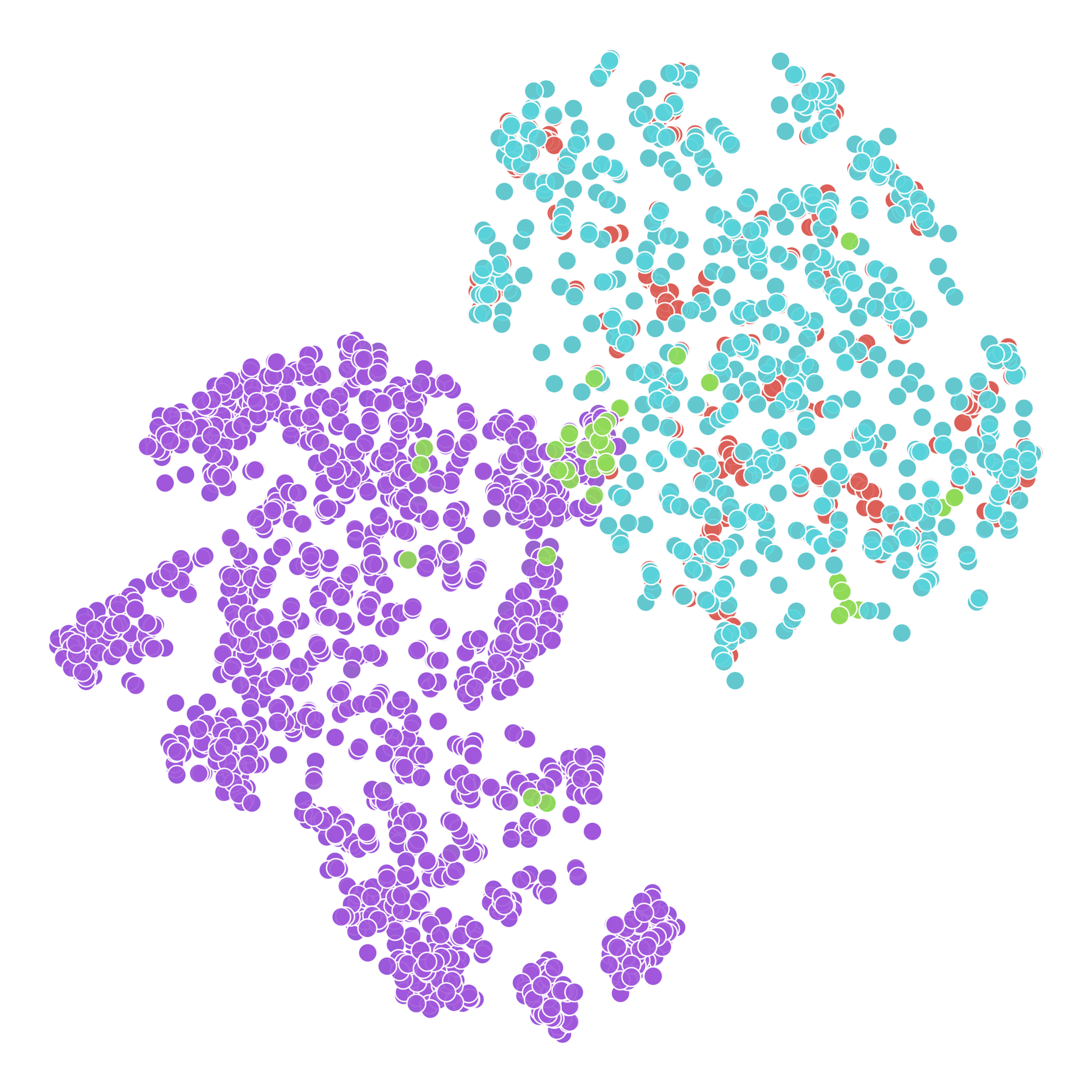}
             \caption{Backbone}
             \label{fig:dann_tsne_bb}
         \end{subfigure}
    \end{center}
    \caption{t-SNE plot of the \CLS{} token extracted from DD branch (a) and backbone (b) for damaged and non-damaged bike instances (BL + Real setting). \textcolor{snsazzurro}{\textbullet{}} Synthetic no damage \textcolor{snsred}{\textbullet} Synthetic damaged \textcolor{snsviola}{\textbullet} Real no damage \textcolor{snsgreen}{\textbullet} Real damaged }
    \label{fig:tsne_dmg_bb}
\end{figure}

The t-SNE plots in \autoref{fig:tsne_dmg_bb} illustrate the distribution of the \CLS{} tokens from the DD branch and the backbone. By comparing synthetic damaged (red) and undamaged (cyan) examples, it can be seen how the backbone captures features related to the bike model and invariant to the presence of damage, whereas the DD branch clearly distinguishes damaged vs. non-damaged bikes. 

\begin{figure}[tb]
    \centering
     \begin{subfigure}[b]{0.3\textwidth}
         \centering
         \includegraphics[width=\textwidth]{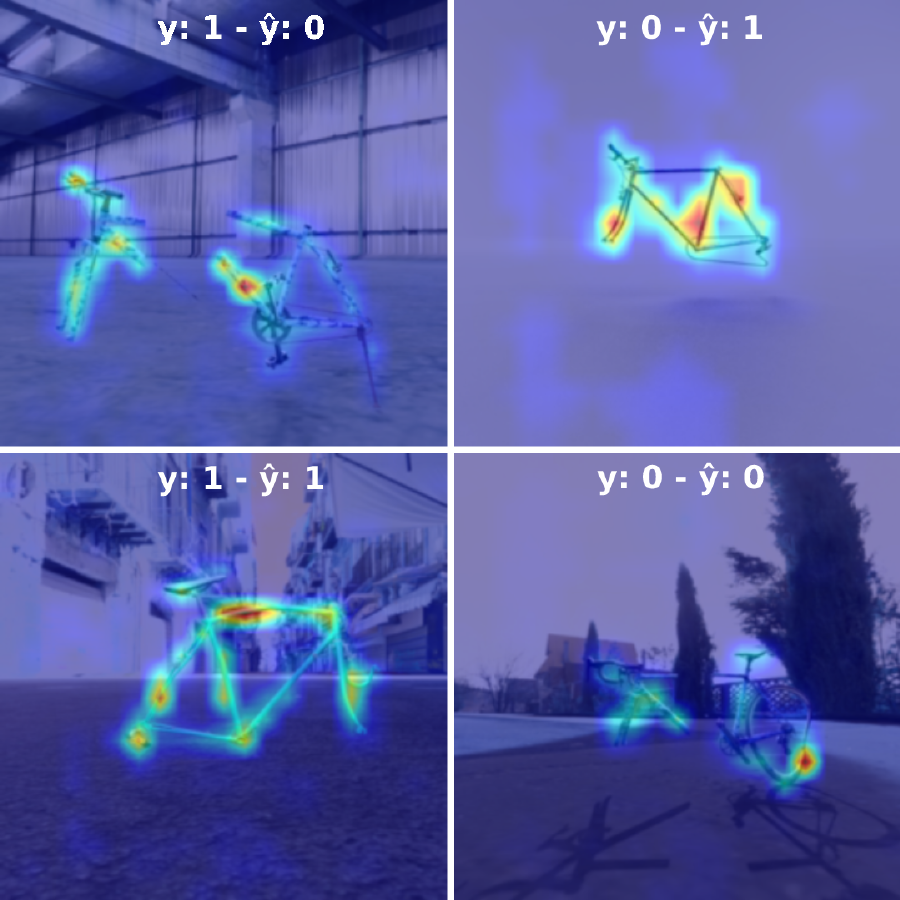}
         \caption{BL + Real}
         \label{fig:xai_bl}
     \end{subfigure}
     \hfill
     \begin{subfigure}[b]{0.3\textwidth}
         \centering
         \includegraphics[width=\textwidth]{images/results/DANN.pdf}
         \caption{BL + Real + DANN}
         \label{fig:xai_dann}
     \end{subfigure}
     \hfill
     \begin{subfigure}[b]{0.3\textwidth}
         \centering
         \includegraphics[width=\textwidth]{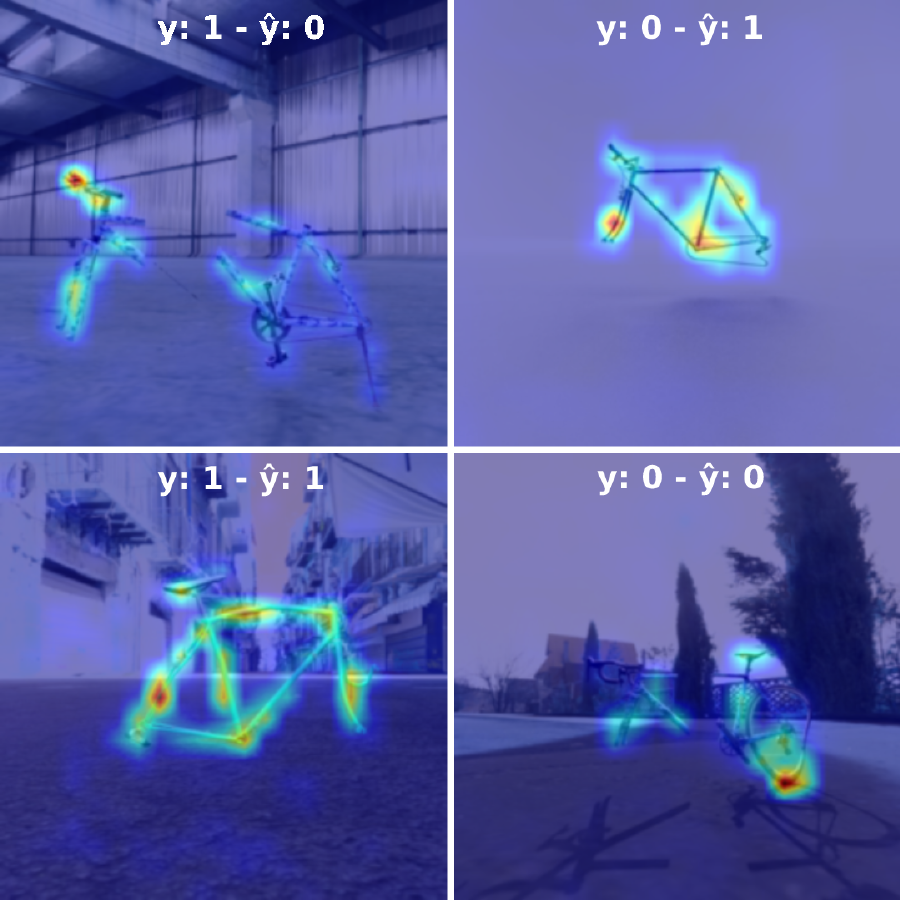}
         \caption{BL + Real + PADA}
         \label{fig:xai_PADA}
     \end{subfigure}
        \caption{Attention maps of \NETWORK{} under different training regimes. Values for Bent frame labels (y) and predictions ($\hat{y}$) are superimposed on the image.  }
        \label{fig:xai}
\end{figure}

\end{document}